\documentclass[10pt,twocolumn,letterpaper]{article}

\usepackage{cvpr}              %

\usepackage{graphicx}
\usepackage{amsmath}
\usepackage{amssymb}
\usepackage{booktabs}

\usepackage{multirow}
\usepackage[]{xcolor}
\usepackage{amsmath}
\usepackage{float} 
\usepackage{amsfonts}
\usepackage{caption}
\usepackage{subcaption}
\usepackage{enumitem}
\usepackage{array}
\usepackage[accsupp]{axessibility}

\usepackage[pagebackref,breaklinks,colorlinks]{hyperref}

\usepackage[capitalize]{cleveref}
\crefname{section}{Sec.}{Secs.}
\Crefname{section}{Section}{Sections}
\Crefname{table}{Table}{Tables}
\crefname{table}{Tab.}{Tabs.}

\def\cvprPaperID{7763} %
\def\confName{CVPR}
\def\confYear{2023}

\definecolor{amethyst}{rgb}{0.6, 0.4, 0.8}
\definecolor{darkpastelgreen}{rgb}{0.01, 0.75, 0.24}
\definecolor{amber}{rgb}{1.0, 0.75, 0.0}
\definecolor{cadmiumorange}{rgb}{0.93, 0.53, 0.18}
\definecolor{lawngreen}{rgb}{0.49, 0.99, 0.0}
\definecolor{limegreen}{rgb}{0.2, 0.8, 0.2}
\definecolor{neongreen}{rgb}{0.22, 0.88, 0.08}
\definecolor{amethyst}{rgb}{0.6, 0.4, 0.8}
\definecolor{darkpastelgreen}{rgb}{0.01, 0.75, 0.24}
\definecolor{greenbest}{RGB}{88,137,15}
\definecolor{redworst}{RGB}{137,15,27}

\definecolor{redpaper}{RGB}{196,77,88}
\definecolor{greenpaper}{RGB}{88,137,15}

\newcommand{\REMOVE}[1]{{}}

\newcommand{\Elena}[1]{\textcolor{cadmiumorange}{[\textbf{Elena}: {#1}]}}

\newcommand{\David}[1]{\textcolor{amethyst}{[\textbf{David}: {#1}]}}

\newcommand{\inputimage}{\text{X}}
\newcommand{\map}{\text{M}}
\newcommand{\mapstack}{\textbf{M}}

\newcommand{\svbrdf}{{\small{SVBRDF}}}
\newcommand{\svbrdfs}{{\small{SVBRDFs}}}
\newcommand{\umetric}{{\sigma_{\text{BRDF}}}}
\newcommand{\unormals}{\sigma_{\measuredangle}}
\newcommand{\urough}{\sigma_{\text{rough}}}
\newcommand{\uspec}{\sigma_{\text{spec}}}

\newcommand{\GreenColor}[1]{\textcolor{greenpaper}{\textbf{#1}}}
\newcommand{\RedColor}[1]{\textcolor{redpaper}{\textbf{#1}}}

\begin{document}

\title{UMat: Uncertainty-Aware Single Image High Resolution Material Capture}

\author{Carlos Rodriguez-Pardo$^{1,2}$~~Henar Dominguez-Elvira$^{1,2}$~~David Pascual-Hernandez$^{1}$~~Elena Garces$^{1,2}$\\[0.2cm]
$^1$SEDDI, Spain ~~ $^2$Universidad Rey Juan Carlos, Spain \\[0.1cm]
}

\maketitle
\REMOVE{
\twocolumn[{%
	\renewcommand\twocolumn[1][]{#1}%
	<\maketitle
	\begin{center}
		\centering
		\captionsetup{type=figure}
		\includegraphics[width=\textwidth]{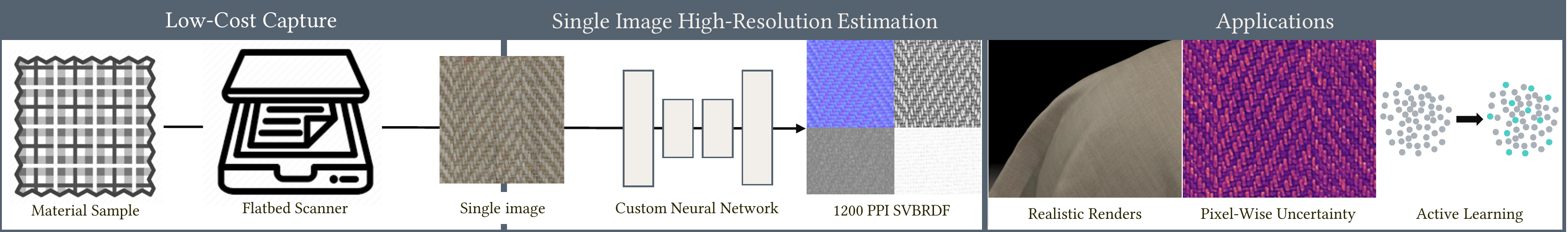}
		\vspace{-5mm}
		\captionof{figure}{\label{fig:teaser}\emph{Textura} is a single-image high-resolution material digitization system which leverages flatbed scanners and deep learning. \Elena{meteria teaser de single column} }
		\vspace{-2mm}
	\end{center}%
}]
}
\begin{abstract}
\REMOVE{
We present \emph{Textura}, a  learning based capture system which can recover high-resolution microgeometry details of materials using a single image captured with a flatbed scanner. Previous material capture systems either rely on inaccessible devices for accurate results, or data-driven approaches for coarser but cheaper digitizations, typically through a single flash-lit image or multiple images captured with a smartphone. In contrast, we show that single images captured with commodity flatbed scanners, and a purposely designed neural network, can bring the best of both worlds: accessible yet accurate material capture. We further propose local uncertainty estimations in a perceptual space, which strongly correlates with test-time error, and may also be used for local artifact detection. We introduce a set of evaluation metrics which may be used for comparing material estimation models in more accurate ways than simple pixel-wise norms. Through a series of ablation studies, we validate our model design and show that we can estimate microgeometry at higher resolutions than previous work.
}

We propose a learning-based method to recover normals, specularity, and roughness from a single diffuse image of a material, using microgeometry appearance as our primary cue. 
Previous methods that work on single images tend to produce over-smooth outputs with artifacts, operate at limited resolution, or train one model per class with little room for generalization. 
In contrast, in this work, we propose a novel capture approach that leverages a generative network with attention and a U-Net discriminator, which shows outstanding performance integrating global information at reduced computational complexity.
We showcase the performance of our method with a real dataset of digitized textile materials and show that a commodity flatbed scanner can produce the type of diffuse illumination required as input to our method.
Additionally, because the problem might be ill-posed --more than a single diffuse image might be needed to disambiguate the specular reflection-- or because the training dataset is not representative enough of the real distribution, we propose a novel framework to quantify the model's confidence about its prediction at test time. 
Our method is the first one to deal with the problem of modeling uncertainty in material digitization, increasing the trustworthiness of the process and enabling more intelligent strategies for dataset creation, as we demonstrate with an active learning experiment.

\end{abstract}

\section{Introduction}\label{sec:introduction}

\begin{figure}[htb!]
	\centering
	\includegraphics[width=\columnwidth]{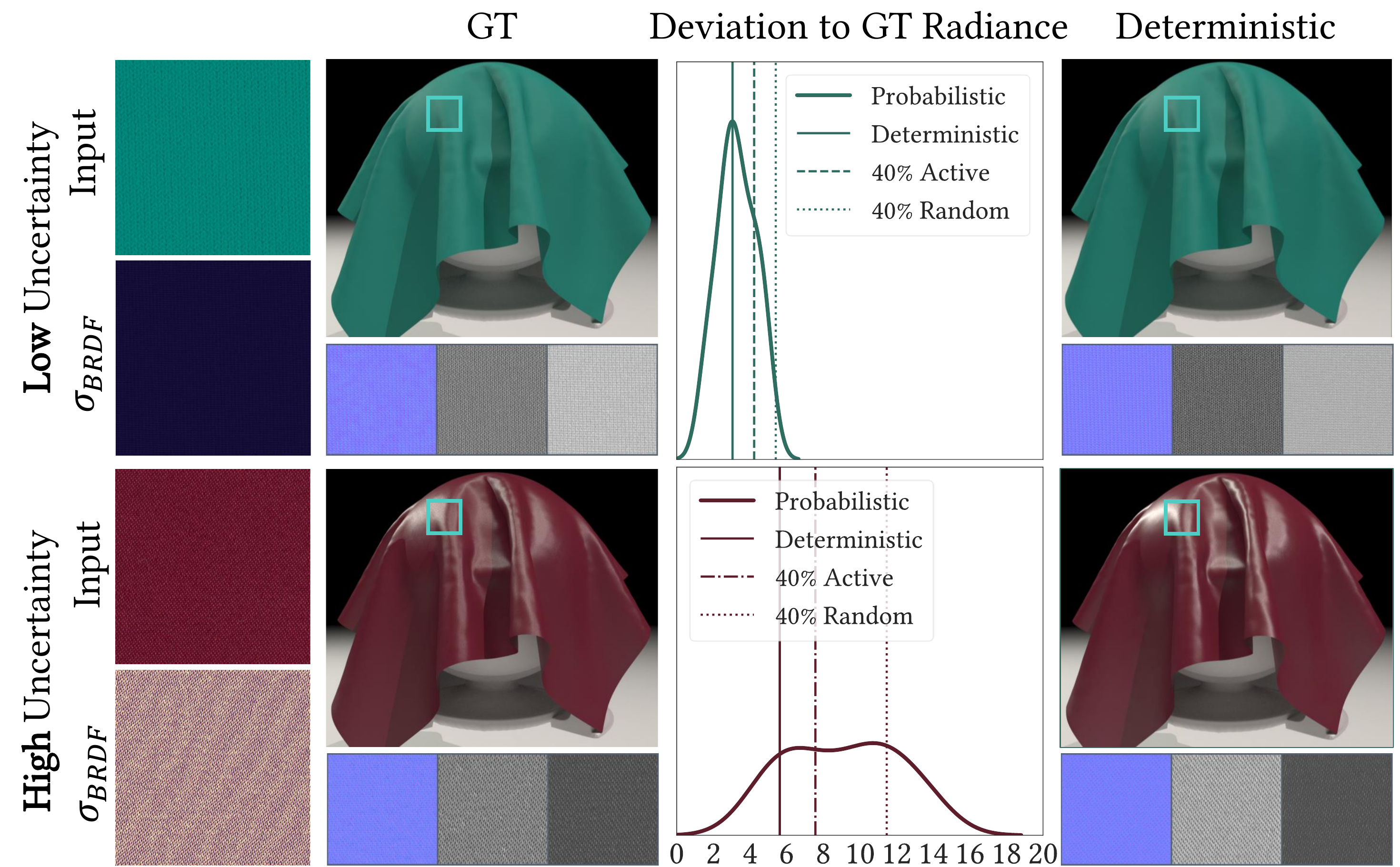}
	\vspace{-5mm}
	\caption{Our method digitizes a material taking as input a single scanned image. Further, it returns a pixel-wise metric of uncertainty $\umetric$, computed at test time through probabilistic sampling, proven useful for active learning. 
	In the plot we compare the average deviations of the radiance of different renders in the blue crop w.r.t the ground truth (GT) of: 1) the distribution of the probabilistic samples of a model trained with 100\% of the data; 2) the deterministic output of that model; 3) the output of a model trained using 40\% of the training dataset, sampled by active learning guided by $\umetric$ and; 4) a model trained using 40\% of the training dataset, randomly sampled.
	The material at the bottom, for which the model shows a higher uncertainty, generates more varied renders and differs most from the ground truth.
	}
	\vspace{-6mm}
	\label{fig:teaser}
\end{figure}

Virtual design, online marketplaces, product lifecycle workflows, AR/VR, videogames, \ldots,  all require lifelike digital representations of real-world materials (\emph{i.e.}, digital twins).
Acquiring these digital copies is typically a cumbersome and slow process that requires expensive machines and several manual steps, creating roadblocks for scalability, repeatability, and consistency. 
Among the many industries requiring digital twins of materials, the fashion industry is in a critical position; facing the demand to digitize hundreds of samples of textiles in short periods, which cannot be achieved with the current technology.

In this context, casual capture systems for optical digitization provide a promising path for scalability. These systems leverage handheld devices (such as smartphones), one or more different illuminations, and learning-based priors to estimate the material's diffuse and specular reflection lobes.
However, existing approaches present several drawbacks that make them unsuitable for practical digitization workflows. 
Generative solutions~\cite{guo2020materialgan,vecchio2021surfacenet} typically produce unrealistic artifacts. 
Despite recent attempts to improve tileability and controllability~\cite{zhou2022tilegen}, these solutions are slow to train and to evaluate (requiring online optimization iterations), are limited in resolution, 
and present challenges for generalization (requiring one model per material class). 
Further, the fact that these methods build on perceptual losses --not pixel losses-- to compare the input photo with the generated material entails extra difficulties when it comes to guaranteeing the repeatability and consistency required for building a digital inventory (i.e., color swatches, prints, or other variations).
On the other hand, methods that build on differentiable node graphs~\cite{henzler2021neuralmaterial} overcome the tileability and resolution limitations, yet, they share the problems derived from using perceptual losses and category-specific training.

\begin{figure}[tb!]
	\centering
		\vspace{-2mm}
	\includegraphics[width=\columnwidth]{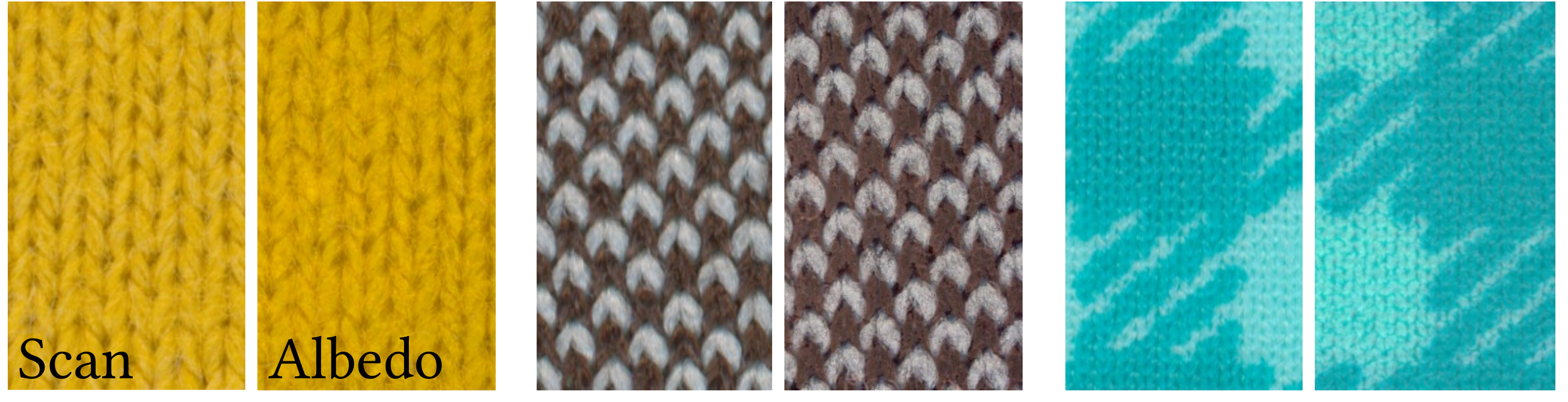}
	\vspace{-6mm}
	\caption{Scanner images vs fitted albedos.
	}
	\vspace{-7mm}
	\label{fig:scanner_vs_albedos}
\end{figure}

In this work, we present \emph{UMat}, a practical, scalable, and reliable approach to digitizing the optical appearance of textile material samples using \svbrdfs. Commonly used \svbrdfs~typically contain two reflection terms: a diffuse term, parameterized by an albedo image, and a specular one, parameterized by normals, specularity, and roughness. Prior work typically estimates both components, which becomes a very challenging problem, obtaining over-smooth outputs, and being prone to artifacts~\cite{deschaintre2018single,guo2021highlight,zhou2021adversarial}. Instead, in this paper, we demonstrate that it is possible to provide accurate digitizations of materials leveraging as input a single diffuse image that acts as albedo and estimating the specular components using a neural network. 
Our key observation is to realize that most of the appearance variability of textile materials is due to its microgeometry and that a commodity flatbed scanner can approximate the type of diffuse illumination that we require for the majority of textile materials (see Figure~\ref{fig:scanner_vs_albedos}).

Nevertheless, single-image material estimation is still an ill-posed problem in our setting, as reflectance properties may not be directly observable from a single diffuse image. To account for these non-directly observable properties, we propose a novel way to measure the model's confidence about its prediction at test time. 
Leveraging Monte Carlo (MC) Dropout~\cite{gal2016dropout}, we propose an uncertainty metric computed as the variance of sampling and evaluating multiple estimations for a single input in a render space.  
We show that this confidence directly correlates with the accuracy of the digitization, which helps identify ambiguous inputs, out-of-distribution samples, or under-represented classes.
Besides increasing the trustworthiness of the capture process, our confidence quantification enables smarter strategies for dataset creation, as we demonstrate with an active learning experiment.

We pose the estimation as an Image-to-Image Translation problem (I2IT) that directly regresses roughness, specular, and normals, from a single input image.
Under the hood, our novel residual architecture has a single encoder enhanced with lightweight attention modules~\cite{wang2020linformer, mehta2021mobilevit} for improving global consistency and reducing artifacts, specialized decoders for each target reflectance map, and a U-Net discriminator~\cite{schonfeld2020u}, which enhances generalization.

\REMOVE{
Posing the problem as an I2IT regression approach has several advantages in terms of simplicity in the learning methodology, however, it also presents several challenges as results tend to be oversmooth and are prone to contain artifacts~\cite{deschaintre2018single,guo2021highlight,zhou2021adversarial}.
We overcome these issues by proposing a novel deep learning architecture and training methodology that we demonstrate to provide accurate and sharp results without artifacts. Under the hood, our novel residual architecture leverages a single encoder enhanced with lightweight attention modules~\cite{wang2020linformer, mehta2021mobilevit} for improving the model understanding of its inputs and removing artifacts, specialized decoders for each target reflectance map, maximizing sharpness and accuracy; and a U-Net discriminator~\cite{schonfeld2020u}, which enhances generalization.
}

\noindent
In summary, we present the following contributions:
\begin{itemize}[noitemsep,topsep=0pt,parsep=0pt,partopsep=0pt]
	\item A novel material capture system which leverages the diffuse illumination provided by flatbed scanners for high-resolution, scalable and reliable digitizations.
	\item An attention-enhanced GAN model and training procedure designed for maximizing accuracy and sharpness, and removing undesired artifacts.
	\item A generic uncertainty quantification framework for material capture algorithms which correlates with prediction error on a render space.
\end{itemize}

\REMOVE{
\emph{Digital twins}, or lifelike digital representations of real-world materials, are essential for enabling the creation of realistic 3D environments. These play an important role in different industries, such as entertainment, science, engineering, art or fashion. Realistic digital twins captured at scale are needed for these fast-paced industries.  
While causal capture systems have been developed for the digitization of optical appearance of materials, using learning-based approaches and low-cost capture devices such as smartphones, these typically lack the level of quality and predictability which is needed for accurate digital twins. Conversely, specialized material capture devices and algorithms have been proposed. These provide higher-quality digitizations but are more expensive, slower and require highly specialized labor. 

Our goal is to obtain a material digitization system which can provide both higher levels of quality and predictability than previous casual capture methods, and higher levels of scalability than what specific capture devices can achieve. Key to the optical behavior of materials is their \emph{microgeometry}, which plays an important role in their appearance at different scales. Previous learning based methods either provide a coarse representation of this microgeometry, which limits their quality; or require multiple images to approximate it, which limit their scalability. 

We propose the use of commodity \emph{flatbed scanners} as a material capture device, because they provide very-high resolution images with constant (albeit unknown) camera and illumination configurations. Estimating the microgeometry and optical properties of a material from a single image captured with a flatbed scanner is a highly ill-posed problem. To approximate it, we leverage a dataset of materials captured at a high resolution, a purposely-designed neural network, and a specific policy of data augmentation. 

The performance of deep-learning based systems is highly dependent on the quality of their inputs, as well as the model design, and the dataset in which the models were trained on. Even if our proposed system captures material properties at a higher resolutions than previous causal methods, it may lack the predictability than specific capture devices may provide. To overcome this issue, we propose a test-time uncertainty estimation method which provides local and perceptually-guided measures of how confident the system is with respect to its own predictions. 
By an extensive set of ablation studies, specific metrics and evaluations, we validate that our method can capture the microgeometry and overall optical appearance of materials from single images captured with a flatbeds scanner, using an attention-guided neural network, providing a new research direction for predictable high-quality material digitizations at scale.

In summary, we propose the following contributions:
\begin{itemize}
    \item The first end-to-end pipeline capable of capturing SVBRDFs using a single image from a commodity flatbed scanner.
    \item A model architecture and training framework designed for generating highly detailed outputs while achieving a global understanding of the input images at a low parameter count.
    \item A test-time perceptual-aware uncertainty estimation method which provides accurate pixel-wise error estimates at test time.
   
    \item An extensive model evaluation framework which helps understand the perceptual influence of the accuracy of the estimated maps.
\end{itemize}

}

\section{Related Work}\label{sec:related_work}

\REMOVE{
\begin{figure*}[tb!]
	\centering
	\includegraphics[width=\textwidth]{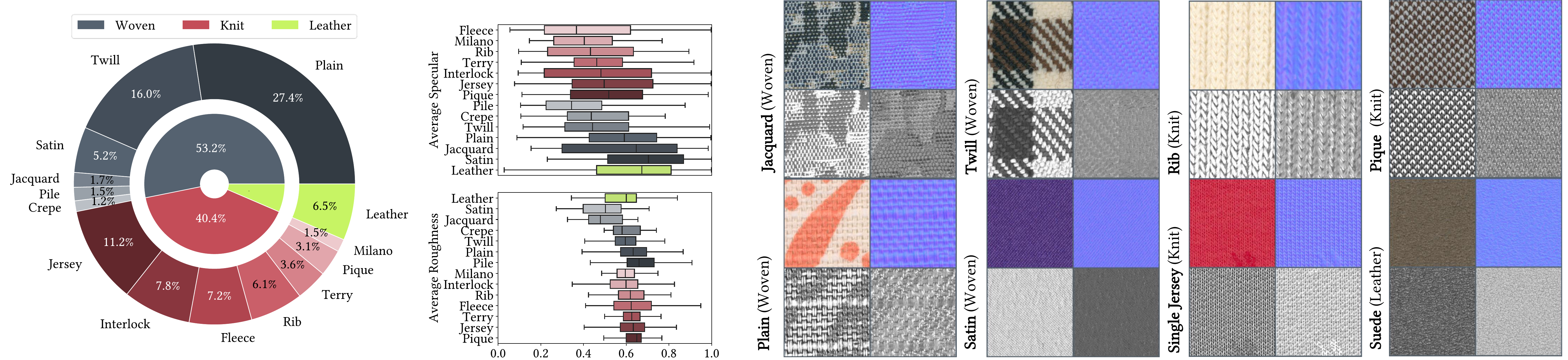}
	\vspace{-5mm}
	\caption{Visualization of our dataset. On the left, we show the percentages of materials in our training dataset, including more detailed subcategories. On their right, we show the average specular and roughness for every category, ordered by family and category. We include some visualizations of 1x1 cm SVBDRF samples, showcasing the variability of microstructures present in these materials.
	}
	\vspace{-5mm}
	\label{fig:EDA}
\end{figure*}}

\begin{figure*}[htb!]
	\centering
	\vspace{-5mm}
	\includegraphics[width=\textwidth]{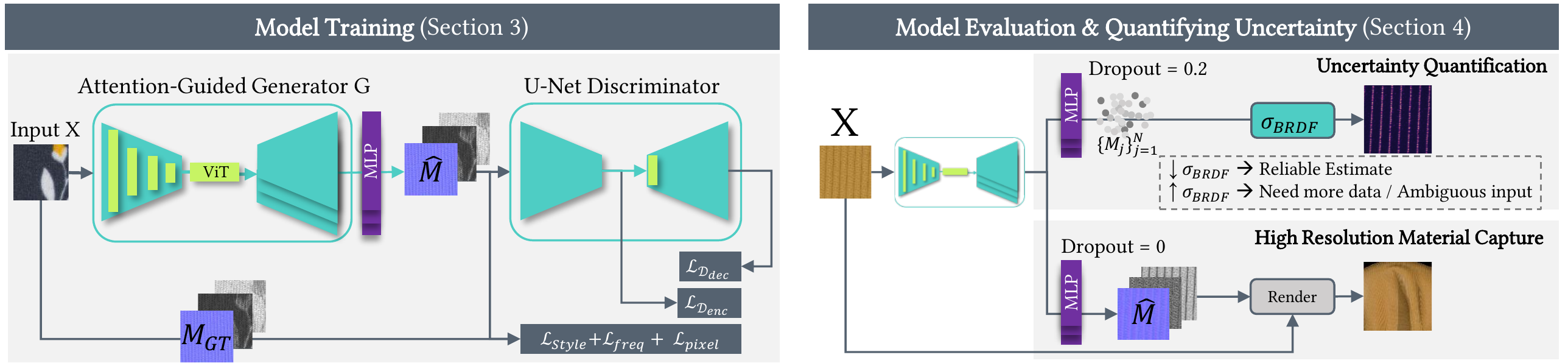}
	\vspace{-7mm}
	\caption{Overview of \emph{UMat}. We propose an attention-guided generator with a U-Net discriminator trained with style, frequency, pixel-wise, and adversarial losses. In green, we show the components that include any form of attention mechanism. The supplementary material contains the detailed architectures. On the right, we show two applications of our method: First, thanks to our test-time uncertainty evaluation, we can provide a measure of reliability of the estimation. Second, the maps that we produce can be used by any render engine. 
 }
	\vspace{-5mm}
	\label{fig:overview_large}
\end{figure*}

\paragraph*{Single Image SVBRDF Capture} Capturing accurate \svbrdfs~from a single image is a challenging problem that requires predicting the photometric response of a material given only a sample of it. The most common approximation is to use a flash-lit front planar image captured with a smartphone.  %
Extending neural style transfer~\cite{gatys2015neural} to material capture, Aittala~\etal~\cite{aittala2016reflectance} leverage pre-trained CNNs and texture priors for smartphone material acquisition. Relatedly, Henzler~\etal~\cite{henzler2021neuralmaterial} use style losses for training generative models for BRDF synthesis. These approaches are optimized for synthesis, which allow for seamlessly tileable outputs, and do not require supervised training. However, they work best for stochastic materials, limiting their scope.

Leveraging datasets of labeled materials, different methods have trained autoencoders for \svbrdf~capture. Li~\etal~\cite{li2017modeling} reconstruct spatially-varying albedo and normals using a U-Net~\cite{ronneberger2015u} and homogeneous specular albedo and roughness using a CNN regressor. This work was extended through Self-Augmented CNNs in~\cite{ye2018single}. By leveraging CRFs, a material classifier and one decoder per map, Li~\etal~\cite{li2018materials} reconstruct spatially-varying albedo, roughness and normals. Deschaintre~\etal~\cite{deschaintre2018single} propose a modified U-Net, synthetic datasets and a render loss. Cascaded models~\cite{li2018learning,sang2020single}; and deep latent spaces optimized using inverse rendering~\cite{gao2019deep} have also shown success for this problem. Recently, Generative Adversarial Networks (GANs) have shown improved capabilities compared to more naive losses. These require a discriminator, which can be trained on renders~\cite{wen2022svbrdf}, \svbrdf~maps~\cite{guo2021highlight,vecchio2021surfacenet}, or both~\cite{zhou2021adversarial}. 

Our approach differs from previous work in several factors. Importantly, we use flatbed scanners instead of smartphones for material capture. While they limit the materials which can be captured, they provide adequate illumination for easier digitizations, and a higher level of resolution and detail. %
We hypothesize that material specularity can be estimated accurately by leveraging its microgeometry. From this assumption, we build a GAN which, in contrast with previous work, leverages state-of-the-art attention mechanisms and discriminator design for obtaining a more holistic understanding of its inputs. Further, we train exclusively on real data and propose a more comprehensive evaluation.%

\REMOVE{
\paragraph*{Multiple Image SVBRDF Capture}
A different corpus of work relies on multiple images of the material%
. By capturing more evidence of the material photometric response, they provide more accurate SVBRDFs. This hinders scalability, as they require a larger capture and calibration effort. The simplest approach is to require two samples, a diffuse and a flash-lit image~\cite{aittala2015two, boss2020two}. More flexible alternatives allow for more samples, combined with a learned prior~\cite{guo2020materialgan, deschaintre2019flexible}; or monte carlo rendering~\cite{luan2021unified}. A different approach is to capture the material at a high resolution, and transfer those details to larger samples of the same material~\cite{deschaintre2020guided,rodriguezpardo2021transfer}. Finally, videos for material acquisition have also shown to be promising capture systems~\cite{ye2021deep}.}

Procedural graphs have also been used for material generation. Instead of relying on material priors, these approaches work by optimizing a material graph through a differentiable pipeline. These provide interesting capabilities, such as easy edition or tiling, but are limited by the expressiveness of the procedural model. These have been explored for general \svbrdf~estimation~\cite{shi2020match, hu2022inverse, guo2020bayesian}, or for high-quality woven fabric digitizations~\cite{jin2022woven}.

\paragraph*{Uncertainty Quantification in Deep Learning}
Measuring the confidence of deep learning models is an active research area~\cite{kendall2017uncertainties} with multiple applications, including safety-critical problems, like self-driving~\cite{huang2019evaluation} or medicine~\cite{kurz2022uncertainty}; and active dataset creation~\cite{lei2021active,soleimany2021evidential}. In computer vision, uncertainty quantification has focused on image classification~\cite{sensoy2018evidential,lei2021active}, segmentation~\cite{krygier2021quantifying}, and depth regression~\cite{huuncertainty,amini2020deep}. To overcome the computational intractability of Bayesian Neural Networks, %
different approximations have been proposed, including MC Dropout~\cite{gal2016dropout}, Deep Ensembles~\cite{lakshminarayanan2017simple} or Variational Inference~\cite{mescheder2017adversarial}. Orthogonal alternatives exist, including Evidential Deep Learning~\cite{amini2020deep, sensoy2018evidential} or frequentist approaches like Constrained Ordinal Regression~\cite{huuncertainty}. We refer the reader to recent surveys~\cite{jospin2022hands,gawlikowski2021survey} for more comprehensive reviews. 

Quantifying uncertainties allows to communicate the end users %
that the model predictions may be inaccurate, suggesting alternative pathways; as well as cheaper dataset creation through~\emph{active learning}. Bayesian material parameter estimation has been proposed for procedural frameworks~\cite{guo2020bayesian}, but, to the best of our knowledge, it has not been explored for \svbrdf~estimation. We aim to propose an efficient uncertainty quantification framework for deep \svbrdf~capture methods which accounts for material perception, and is agnostic to the material model. %

\section{Method}\label{sec:overview}

Our method starts from an input image $\inputimage$ of a material taken under diffuse lighting and outputs the parameters of the specular lobe of the \svbrdf, \textit{i.e.}, $\mapstack = \{ \map_i\}_{i=1}^{3}$ corresponding to the material roughness, specularity, and normals. We illustrate this process in Figure~\ref{fig:overview_large}. Following previous work~\cite{karis2013real,munkberg2022extracting}, we use the physically-based material model from Disney~\cite{burley2012physically}, which aggregates a diffuse term with an isotropic, microfacet specular GGX lobe $s(\mapstack)$~\cite{walter2007microfacet}, such that,
$
f_{l,v}(\mapstack,\inputimage) = \frac{\inputimage}{\pi} + s_{l, v}(\mapstack) \in \mathbb{R}^{x \times y}
$
is the shading model for a light $l$ and view position $v$. 
We formulate this estimation as an Image-to-Image Translation problem (I2IT). We train a U-Net~\cite{ronneberger2015u} generator $G(\inputimage)=\hat{\mapstack}$ within a GAN framework, extending the adversarial loss with pixel-wise, style, and frequency losses. We design the generator to maximize accuracy, sharpness, and robustness. To do so, we train a single encoder, enhanced with self-attention layers and a transformer, and use one decoder per map.  Sections~\ref{sec:netdesign},~\ref{sec:loss}, and~\ref{sec:dataaugment} present the network design, the loss and data augmentation choices, respectively. Implementation details are provided on the supplementary material.

Using as input a single image taken under diffuse lighting presents extra challenges when estimating the \svbrdf; we lack the extra cues provided by more complex illumination patterns (e.g. flash lighting). Therefore, in Section~\ref{sec:uncertainty_description}, we explain how to compensate this potential ambiguity by introducing an uncertainty metric that can be computed at test time.  Section~\ref{sec:evaluation} presents the evaluation, which includes the description of our dataset and metrics (Sections~\ref{sec:dataset} and~\ref{sec:metrics}), an ablation study that validates the design (\ref{sec:ablation}), qualitative and quantitative results (\ref{sec:qualitative} and~\ref{sec:uncertainty}), an application of our uncertainty metric in an active learning setting (\ref{sec:activelearning}), and comparisons with previous work (\ref{sec:comparisons}).

\REMOVE{
\begin{figure}[tb!]
	\centering
	\includegraphics[width=\columnwidth]{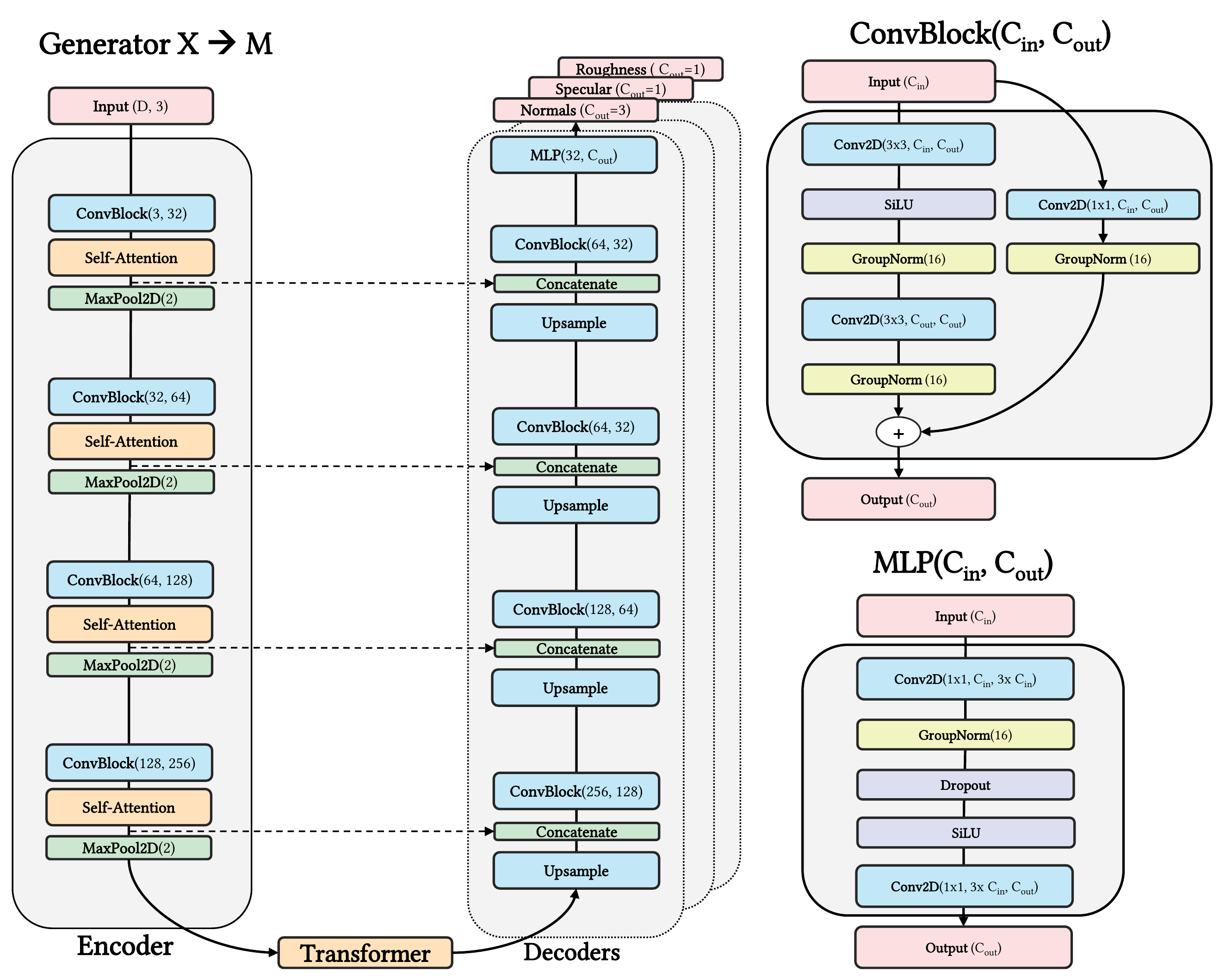}
	\caption{An overview of our generator architecture. It contains a single encoder, extended with self-attention layers with linear complexity. We extend the bottleneck with a lightweight ViT. Each map is estimated with a different decoder, which is ended with a pixel-wise MLP. Every upsampling layer is learned through transposed convolutions, and we employ residual connections, group normalization and SiLU non-linearities. The number of input and output channels for each layer are marked as $(C_{in},C_{out})$, while spatial dimensions are specified as $D$. 
	}
	\vspace{-5mm}
	\label{fig:generator_architecture}
\end{figure}
}

\subsection{Network Design} \label{sec:netdesign}

We use a U-Net~\cite{ronneberger2015u} with residual connections~\cite{he2016deep,diakogiannis2020resunet,deschaintre2021deep} in all of our convolutional blocks, an individual decoder per map~\cite{garces2022survey,rodriguezpardo2022seamlessgan,deschaintre2021deep,zhou2021adversarial}, and group normalization~\cite{wu2018group}. %
To each decoder, we add a pixel-wise dropout-regularized MLP~\cite{srivastava2014dropout}, aimed at increasing the accuracy of the predictions and to allow us to measure uncertainty at test-time. 

\noindent
While this multi-decoder residual U-Net is relatively accurate, it is limited by its receptive field, as is common on fully-convolutional architectures. Previous work~\cite{deschaintre2018single} proposed the use of a \emph{global track} for fusing spatially distant information. We instead draw inspiration from recent advances in attention and diffusion models~\cite{saharia2022palette,ho2020denoising,rombach2022high}, and add a self-attention module with linear complexity~\cite{wang2020linformer} to the output of every convolutional block in the encoder. 
Finally, we add a lightweight MobileViT~\cite{mehta2021mobilevit} to the bottleneck to provide the model with a global understanding of its input.
By performing the most complex computations at the encoder, we provide the specialized decoders with dense inputs which are computed only once.

\subsection{Loss Function} \label{sec:loss}

Our loss function is comprised of four terms: pixel-wise losses, an adversarial loss, a style loss, and a frequency loss:
\begin{equation}
	\mathcal{L}_{G} = \sum_{i} \lambda_i \mathcal{L}_{pixel_i} + \lambda_{adv}  \mathcal{L}_{adv}  + \lambda_{style}   \mathcal{L}_{style} + \lambda_{freq}   \mathcal{L}_{freq}
\end{equation}
$\mathcal{L}_{pixel}$ is the $\mathcal{L}_1$ norm weighted per map, $\lambda_i$. $\mathcal{L}_1$ produces sharper results than higher-order alternatives, such as $\mathcal{L}_2$. We introduce an adversarial loss to handle the intrinsic ambiguity of ill-posed problems~\cite{texler2020interactive,isola2017image,garces2022survey,vecchio2021surfacenet}. In our case, the choice of the discriminator is a critical design decision. 
Recent work~\cite{schonfeld2020u} proposed U-Net architectures for discriminators, which allows to better learn both low and high-level features, and to introduce further regularization. These result in more conservative albeit less diverse generations~\cite{han2022rarity}. We use a U-Net discriminator~\cite{schonfeld2020u} with attention~\cite{woo2018cbam}, which outputs two estimations: a scalar output $\mathcal{D}_{enc}$, provided by its encoder, and a 2D estimation $\mathcal{D}_{dec}$, provided by its decoder. $\mathcal{D}_{enc}$ provides a global estimate of the quality of the stack $\mapstack$, while $\mathcal{D}_{dec}$ gives pixel-wise estimations. As in~\cite{schonfeld2020u}, we add a \emph{regularization} term $\mathcal{L}_{\mathcal{D}_{dec}}^{cons}$, and leverage \emph{cut-mix} as for discriminator data-augmentation. Our discriminator and adversarial losses are:
\begin{align}
	\mathcal{L}_\mathcal{D} &= \mathcal{L}_{{\mathcal{D}_{enc}}} + \mathcal{L}_{\mathcal{D}_{dec}} + \lambda_{cons}\mathcal{L}_{\mathcal{D}_{dec}}^{cons} \\
	\mathcal{L}_{adv} &= \text{log}(\mathcal{D}_{enc}(\text{G}(\inputimage)) +  \text{log}(\mathcal{D}_{dec}(\text{G}(\inputimage))
\end{align}
where the implementation of $\mathcal{L}_{{D_{enc}}}$ and $\mathcal{L}_{{D_{dec}}}$ follows~\cite{schonfeld2020u}.

We further add two losses to improve the accuracy and sharpness of the results: a frequency loss $\mathcal{L}_{freq}$ and a style loss $\mathcal{L}_{style}$. $\mathcal{L}_{freq}$ is estimated by averaging the \emph{focal frequency loss}~\cite{jiang2021focal} computed over each individual channel of $\mapstack$. This is designed to help GANs preserve high-frequency details. Further, as shown in prior work, working in the frequency domain is beneficial when handling textures~\cite{aittala2013practical, mardani2020neural}. 
Our style loss $\mathcal{L}_{style}$ is inspired by the success of neural losses when dealing with textures~\cite{gatys2015neural,gatys2015texture}. However, off-the-shelf metrics which are designed for 3-channel images, are not immediately usable in {\svbrdf}.
While it is possible to compute them for each map separetely~\cite{rodriguezpardo2022seamlessgan}, this does not necessarily preserve inter-map consistency. 
We follow recent work~\cite{chambon2021passing} and use LPIPS~\cite{zhang2018unreasonable} taking as input a 3-channel image created by randomly sampling three channels from the set of five available channels of $\mapstack$.

\subsection{Data Augmentation} \label{sec:dataaugment}

We follow two strategies for data augmentation. First, we perform patch-based training and affine transforms with randomly cropped patches~\cite{rodriguezpardo2021transfer,texler2020interactive,vecchio2021surfacenet}. 
We also apply random rescales for generalization at lower resolutions, and rotations to account for possible misalignments that may arise when capturing the samples. Second, we apply several image transformations to increase model robustness: random intensity changes in HSV space to the inputs, gaussian noise and blurs, and random erasing~\cite{zhong2020random} for regularization.

\section{Uncertainty Estimation}\label{sec:uncertainty_description}

\begin{figure}[tb!]
	\centering
	\vspace{-2mm}
	\includegraphics[width=\columnwidth]{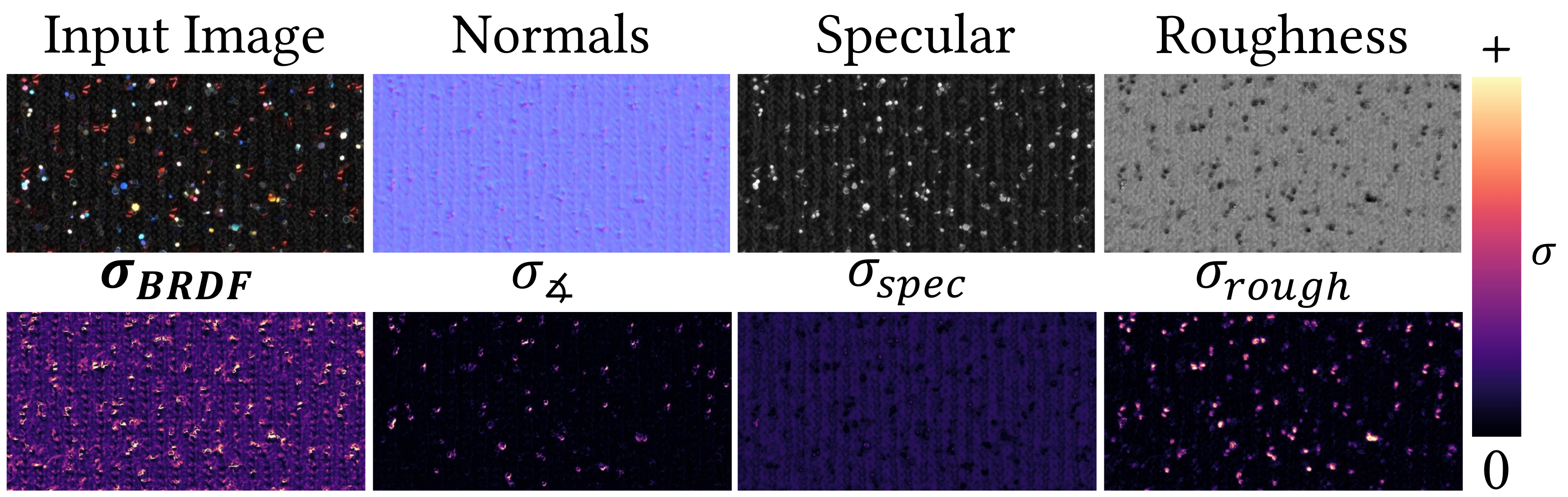}
	\vspace{-7mm}
	\caption{Top: input image of a \emph{rib} material with metallic sequins. Bottom: $\umetric$ and per-map uncertainties.}
	\vspace{-6mm}
	\label{fig:uncertainty}
	
\end{figure}

Material acquisition from a single diffuse image is potentially an ill-posed problem, since it assumes that the microgeometry is a sufficient cue to predict the material appearance. While pure learned priors have been proven to work well for inverse problems of single images object shape estimation~\cite{lichy2021shape,luan2021unified, hwang2022sparse,munkberg2021nvdiffrec,li2018learning}, they are typically combined with render losses that guarantee consistency in the reconstruction. We lack the necessary input to include this kind of supervision, therefore, we propose an alternative approach aimed at quantifying the confidence of the prediction.
This is valuable for several purposes. It provides a way of communicating possible inaccuracies to users of these systems;
and more importantly, it enables for efficient dataset creation through active learning, as we show in Section~\ref{sec:activelearning}.

We propose an uncertainty quantification mechanism which is applied to invidual per-map estimations, and also globally in a render space. 
It is possible to measure uncertainty, among other methods, through deep ensembles~\cite{lakshminarayanan2017simple}, evidential learning~\cite{sensoy2018evidential,bao2021evidential,wang2022uncertainty}, or ordinal regression~\cite{huuncertainty}. However, they are costly to train and evaluate, and would imply major changes in our method.
Instead, we follow a probabilistic approach called MC Dropout~\cite{gal2016dropout}, with which, for a particular input, we sample a set of predictions by adding randomness to the forward pass of our model. This process has no impact in the regular deterministic evaluation and implies no changes in our model architecture.
Specifically, while measuring uncertainty, we randomly deactivate 20\% of the neurons of the MLP of our decoders, obtaining a set of outputs $ \text{U} = \{\hat{\mathbf{\map}}_j \sim \text{G}(\inputimage)\}_{j=1}^{N}$  that we use to compute several metrics.

First, we compute the pixel-wise standard deviation for each invididual map separately, obtaining $\unormals$, $\uspec$, and $\urough$, for the normals, specular, and roughness maps, respectively. 
Then, inspired by perceptual metrics for computing BRDF differences~\cite{lavoue2021perceptual}, we define our novel perceptually-aware uncertainty metric $\umetric$ as follows: 
	\begin{equation} \label{eq:uncertainty}
		\resizebox{\hsize}{!}{$\umetric = \frac{1}{|xy|}\sum\limits_{xy} \log \left( \frac{1}{|S|} \sqrt{\sum\limits_{(l,v) \in S} \sqrt[3]{ \sigma_{l, v} (\{ f_{l,v}(\text{U}_j, K)\cos(\theta_l)\}_{j=1}^{N} )}}\right)$}
	\end{equation}
where $K$ is a 2D image of a constant neutral grey value, $\sigma_{l, v} \in \mathbb{R}^{x\times y}$ is the pixel-wise standard deviation of the renders $f_{l,v}$ obtained for the set of sampled maps U at light $l$ and view position $v$, and $S$ is the set of 50 views optimized in~\cite{nielsen2015optimal} for efficient BRDF capture.
The log spatially-varying result is averaged across the spatial dimensions $xy$ to obtain a single value as output.
This equation introduces perceptual components to our uncertainty metric in two forms: first, by applying cosine weighting to the light position to compensate for light attenuation at grazing angles, and second, by taking the cubic root of these differences to attenuate peak reflectances. Figure~\ref{fig:uncertainty} showcases an example of the per-map and BRDF uncertainties for a rib material with sequins. While the uncertainty is low at the yarns because it is a common material in our dataset, it appears high in the sequins since that effect had not been observed during training.

\vspace{-0.5mm}
\section{Experimental Results}\label{sec:evaluation}

\subsection{Dataset}\label{sec:dataset}

We gathered a novel dataset for training and testing our method. It comprises 2000 textile materials of a variety of families and microstructures that we divided into 14 families: crepe, jacquard, pile, plain, satin, and twill for \textit{wovens}; fleece, french terry, interlock, jersey, milano, pique, and rib for \textit{knits}; and, finally, \textit{leathers}. 
Each family differs in its construction pattern (\emph{i.e.,} its microstructure), which directly impacts its optical appearance. In the supplementary material we show a more detailed analysis of the dataset. For each material, we have an image scanned with the flatbed scanner \textsc{EPSON V850} whose lighting configuration is close to diffuse (as we show in Figure~\ref{fig:scanner_vs_albedos}), and its corresponding ground truth specular maps of the \svbrdf~(normals, specular, and roughness). To obtain these maps, first, we digitized the material with an optical gonioreflectometer and then, propagated the maps to the scan using map propagation techniques~\cite{rodriguezpardo2021transfer}. 
All our images and maps have a resolution of 1000 PPIs, allowing us to leverage the full semantics of the microstructure for inference. We split our dataset in 90-10 for train and test, making sure that every family is equally represented in both splits. %

\REMOVE{We augment our dataset in several ways following well-established strategies~\cite{rodriguezpardo2021transfer,texler2020interactive,vecchio2021surfacenet}, using patch-based training, random rescales, rotations, intensity changes in HSV space, and distorting the input such with gaussian blurs and random erasings~\cite{zhong2020random} for model regularization. }

\subsection{Metrics}\label{sec:metrics}

We quantify individual \textit{per-map accuracy}, \textit{rendered perceptual accuracy}, and \textit{artifacts}. \textbf{Per-map accuracy} is computed differently depending on the semantics of the map: the roughness and specular maps are evaluated using Mean Absolute Error ($\mathcal{L}_1$), and the normals are evaluated using the angular distance in vector space ($\mathcal{L}_{\measuredangle}$). 
Further, to account for the possibility that the model always returns an accurate average value, resulting in relatively low  $\mathcal{L}_1$, we also measure Pearson correlation $\rho$.

We evaluate \textbf{rendered perceptual accuracy} $\mathcal{L}_{\text{\tiny{BRDF}}}$ between the ground truth stack $\mapstack_{GT}$ and the estimation $\hat{\mapstack}$ following existing metrics~\cite{lavoue2021perceptual},
\begin{align} \label{eq:brdf}
	\resizebox{\hsize}{!}{$\mathcal{L}_{\text{\tiny{BRDF}}}= \frac{1}{|xy|} \sum\limits_{xy} \sqrt{ \frac{1}{|S|}\sum\limits_{(l, v) \in S } \sqrt[3]{\cos^2(\theta_l) \left(f_{l,v}(\mapstack_{GT},K) - f_{l,v}(\hat{\mapstack},K)\right)^2}}$}
\end{align}
where the terms are the same as in Equation~\ref{eq:uncertainty}.

\noindent
Finally, we found that some architectural improvements in the neural network introduce artifacts in the specular
and roughness
maps that were not present in the input image, as illustrated in Figure~\ref{fig:ablation_qualitative}. Thus, we provide an \textbf{artifacts detection} metric to quantify them. We start by defining a metric of homogeneity for an input image $\text{I}$, 
\begin{equation}
	\mathcal{H} (\text{I}) = \frac{1}{|xy|}  \sum\limits_{xy} \frac{1}{|d|} \sum_{d = \{ \uparrow, \downarrow, \leftarrow, \rightarrow \}} \| F_{\text{Box}}(\text{I}) - F_{\text{Box}}(\text{I}^{d})  \|_1
\end{equation}
where $\text{I}^{d}$ is the image shifted up, down, left, and right by a number of pixels equal to the kernel size of the box filter. %
Then, we define three metrics that we compute per map: $e_1(\map) = \mathcal{H} (\map)$, $e_2(\map) = \frac{\mathcal{H}(\map)}{\mathcal{H}(\inputimage)}$, and $e_3(\map) =  \text{\emph{MI}}(\inputimage, \map)^{-1}$. 
$\inputimage$ is the original input image, and \emph{MI} is the Mutual Information~\cite{russakoff2004image}, which helps discern artifacts that appear in a single map from semantic patterns (\emph{e.g.,} plaids, prints). 
Each map is labeled as having artifacts if the majority of metrics exceed their corresponding thresholds, $t_m(\map)$. If any of the maps contain artifacts, the entire stack $\mapstack$ is classified as having artifacts. The values for the thresholds and filter are included in the supplementary material.
 \REMOVE{\Elena{move to supp}The thresholds for each material map and the kernel size for the uniformity metric have been optimized given a set of 102 manually labeled textures: $t_1(\map_s) =0.01$, $t_2(\map_s)=1.41$, $t_3(\map_s)=1.33$; $t_1(\map_r) =0.01$, $t_2(\map_r)=0.99$, $t_3(\map_r)=3.12$. The size of the box filter is $s_{\text{Box}}=0.1275\cdot\text{DPI}_{\text{X}}$.
}

\REMOVE{
Our algorithm takes into account three metrics. First, we provide a measure of uniformity,\Elena{any related work here?}\David{not really, it is just what I came up with to measure uniformity between adjacent regions in the images}
\begin{equation}
	U (\text{X}) = \frac{1}{|d|} \sum_{d = \{ \uparrow, \downarrow, \leftarrow, \rightarrow \}} \| f_{\text{Box}}(\text{X}) - f_{\text{Box}}(\text{X}^{d})  \|_1
\end{equation}
where $\text{X}^{d}$ is the image shifted up, down, left, and right by a number of pixels equal to the kernel size of the Box filter. This value is defined in terms of the guidance image resolution so that it always corresponds to the same real-world distance \David{This value is optimized along with the threshold values, as explained in the last paragraph}. We apply this function to both the roughness and the specular maps, obtaining $m_{rough, 1}=U(\text{rough})$, and $m_{spec, 1}=U(\text{spec})$, respectively.

Then, to take into account if the amount of uniformity in the maps inspected is equivalent to the one present in the input image $I'$, we measure their ratio of uniformity as $m_{rough, 2} = \frac{U(\text{rough})}{U(\text{I'})}$ and $m_{spec, 2}=\frac{U(\text{spec})}{U(\text{I'})}$.

Finally, we measure the inverse of the Mutual Information~\cite{russakoff2004image} between each map and the input image $I'$ so that we can discern artifacts from truly irregular patterns, such as those present in tartans or fabrics with prints. In that way, we obtain: $m_{rough,3}=1-\text{MI}(I',\text{rough})$ and $m_{spec,3}=1-\text{MI}(I',\text{spec})$.

If the majority of the metrics $m_n$ defined per map exceed their corresponding thresholds $t_n$, we consider that the map inspected contains artifacts.
\begin{equation}
	\textrm{A}_{map} = \| \left( m_1 > t_1, m_2 > t_2, m_3 > t_3 \right)  \| \geq 2  
\end{equation}
The thresholds for each material map and the kernel size for the uniformity metric have been optimized given a set of 102 manually labeled textures.
}

\subsection{Ablation Study}\label{sec:ablation}

\begin{figure}[tb!]
	\centering
	\includegraphics[width=\columnwidth]{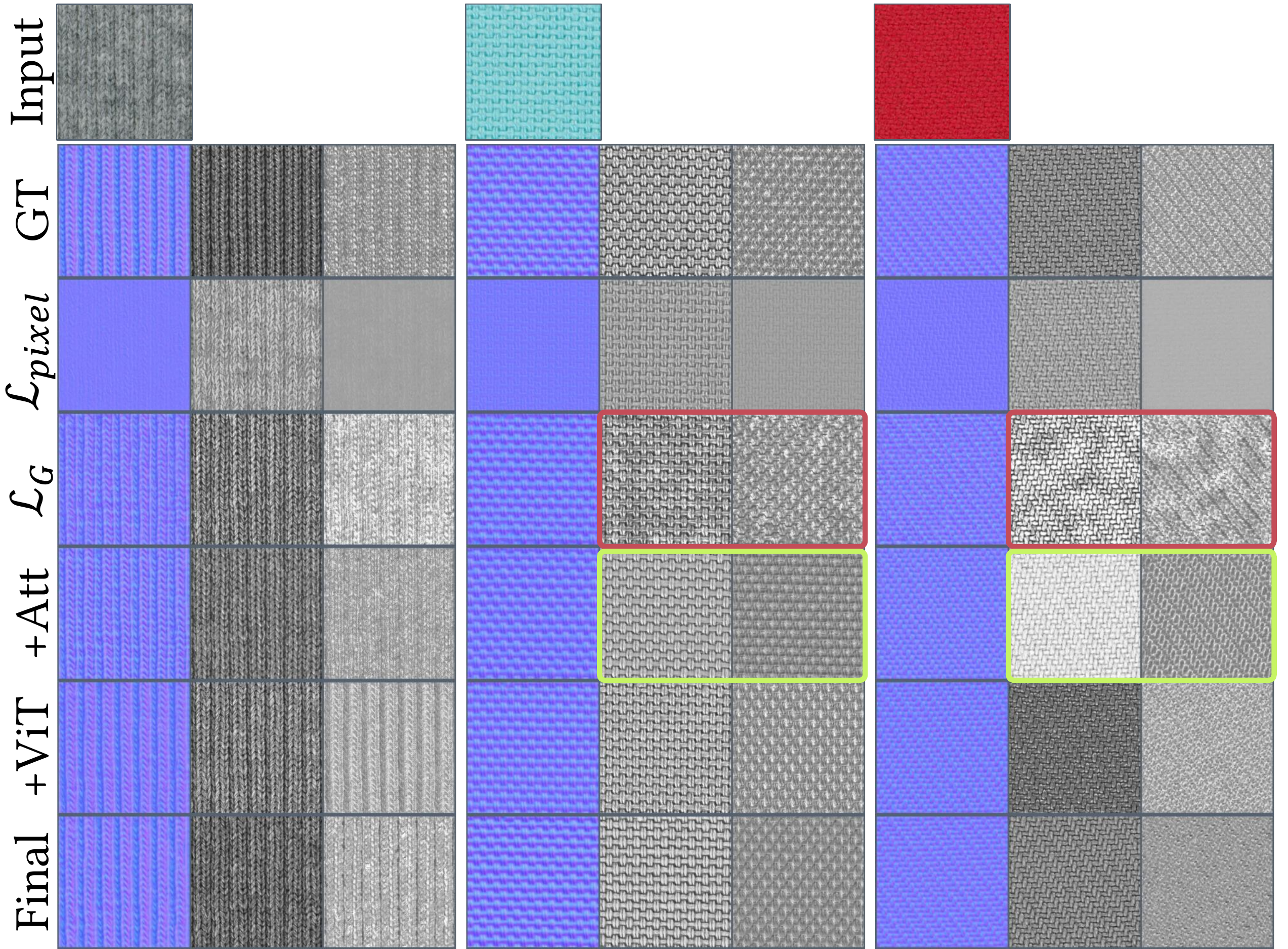}
	\caption{Qualitative results of some configurations of our ablation study. %
		In \RedColor{red}, we show that the baseline generator architecture trained on the full loss introduces artifacts, which are removed using \GreenColor{attention} on the encoder. Further results are included in the supplementary material.
	}
	\label{fig:ablation_qualitative}
\end{figure}

In Figure~\ref{fig:ablation_qualitative} and Table~\ref{tab:ablation-quantitative}, we present an ablation study to validate our model design. From a baseline U-Net~\cite{ronneberger2015u} trained with a pixel-wise loss and no data augmentation other than rescales, we add different components to the model to improve its generalization. First, we observe that using a PatchGAN~\cite{isola2017image,vecchio2021surfacenet} discriminator provides a significant increase in accuracy. However, using a similarly-sized U-Net discriminator~\cite{schonfeld2020u}, we achieve better results, particularly when using discriminator regularization. Further, $\mathcal{L}_{style}$ and $\mathcal{L}_{freq}$ yield significant improvements, most notably in the normal map. To the baseline U-Net trained with the full loss, adding residual connections and one decoder per map increases accuracy. However, this setup tends to produce artifacts as it struggles to integrate global information. By adding self-attention to the encoder, we remove the artifacts; and with the MobileViT~\cite{mehta2021mobilevit}, we achieve higher quality results. Our final model contains full data augmentation, which provides small gains on generalization.

\begin{table}[tb]
	\centering
	\resizebox{\columnwidth}{!}{
		\begin{tabular}{@{}rrrccccccc@{}}
			\cmidrule(l){2-10}
			&
			\multicolumn{2}{c}{\textbf{Configuration}} &
			$\rho^{S}\uparrow$ &
			$\rho^{R}\uparrow$ &
			{$\mathcal{L}_{\measuredangle} \downarrow$} &
			$\mathcal{L}_{1}^{S}\downarrow$&
			$\mathcal{L}_{1}^{R}\downarrow$&
			$\mathcal{L}_{\text{\tiny{BRDF}}}\downarrow$ &
			Art.$\downarrow$ \\ \cmidrule(l){2-10} 
			\multicolumn{3}{r|}{Baseline} &
			
			\RedColor{0.615} &
			\RedColor{0.329} &
			\RedColor{7.570} &
			\RedColor{0.130} &
			\RedColor{0.070} &
			\RedColor{0.325} &
			\GreenColor{\textbf{0.0}} \\ \cmidrule(l){2-10} 
			\multirow{5}{*}{\rotatebox{90}{\textbf{\footnotesize{Loss}}}} &
			Baseline + &
			\multicolumn{1}{r|}{PatchGAN} &
			0.810 &
			0.510 &
			3.729 &
			0.089 &
			0.069 &
			0.307 &
			12.0 \\
			&
			Baseline + &
			\multicolumn{1}{r|}{U-Net D.} &
			0.854 &
			0.658 & 	
			2.950 &
			0.085 &
			0.068 &
			0.299 &
			\RedColor{28.0} \\
			&
			U-Net D. + &
			\multicolumn{1}{r|}{$\mathcal{L}_{\mathcal{D}_{dec}}^{cons}$} &
			0.858 &
			0.653 &
			2.860 &
			0.088 &
			0.068 &
			0.296 &
			19.0 \\
			&
			$\mathcal{L}_{\mathcal{D}_{dec}}^{cons}$ + &
			\multicolumn{1}{r|}{$\mathcal{L}_{style}$} &
			0.831 &
			0.655 &
			2.790 &
			0.091 &
			0.060 &
			0.289 &
			23.0 \\
			&
			$\mathcal{L}_{style}$ + &
			\multicolumn{1}{r|}{$\mathcal{L}_{freq}$} &
			0.856 &
			0.677 &
			2.410 &
			0.086 &
			0.059 &
			0.288 &
			20.1 \\ \cmidrule(l){2-10} 
			\multirow{4}{*}{\textbf{\rotatebox{90}{\footnotesize{Model}}}} &
			$\mathcal{L}_{freq}$ + &
			\multicolumn{1}{r|}{Residual} &
			0.855 &
			0.665 &
			2.310 &
			0.089 &
			0.060 &
			0.285 &
			25.5 \\
			&
			Residual + &
			\multicolumn{1}{r|}{Decoders} &
			0.863 &
			0.699 &
			2.120 &
			0.079 &
			0.054 &
			0.276 &
			18.5 \\
			&
			Decoders + &
			\multicolumn{1}{r|}{{Attention}} &
			0.860 &
			0.665 &
			2.080 &
			0.080 &
			0.059 &
			0.275 &
			0.5 \\
			&
			Atention + &
			\multicolumn{1}{r|}{{ViT}} &
			0.863 &
			0.665 &
			2.040 &
			0.079 &
			0.057 &
			0.271 &
			0.5 \\ \cmidrule(l){2-10} 
			\multirow{4}{*}{\textbf{\rotatebox{90}{\footnotesize{Augment.}}}} &
			ViT + &
			\multicolumn{1}{r|}{Color} &
			\GreenColor{\textbf{0.899}} &
			0.692 &
			1.969 &
			0.068 &
			0.054 &
			0.269 &
			\GreenColor{\textbf{0.0}} \\
			&
			Color + &
			\multicolumn{1}{r|}{Rotations} &
			0.870 &
			0.682 &
			2.050 &
			0.078 &
			0.055 &
			0.271 &
			0.2 \\
			&
			Rotations + &
			\multicolumn{1}{r|}{Distortion} &
			0.876 &
			0.699 &
			2.001 &
			0.074 &
			0.053 &
			0.268 &
			\GreenColor{\textbf{0.0}} \\
			&
			Distortion + &
			\multicolumn{1}{r|}{Erasing} &
			0.893 &
			\GreenColor{\textbf{0.727}} &
			\GreenColor{\textbf{1.941}} &
			\GreenColor{\textbf{0.067}} &
			\GreenColor{\textbf{0.052}} &
			\GreenColor{\textbf{0.265}} &
			\GreenColor{\textbf{0.0}} \\ \cmidrule(l){2-10} 
	\end{tabular}}
	\vspace{-2mm}
	\caption{Results of our ablation study, across a variety of metrics. 
		\emph{Art.} refers to our artifact detection metric. 
		We use a color code to highlight \GreenColor{best} and \RedColor{worst} cases.}
	\label{tab:ablation-quantitative}
	\vspace{-4mm}
\end{table}

\subsection{Qualitative Analysis}\label{sec:qualitative}
\begin{figure}[tb!]
	\centering
	\includegraphics[width=\columnwidth]{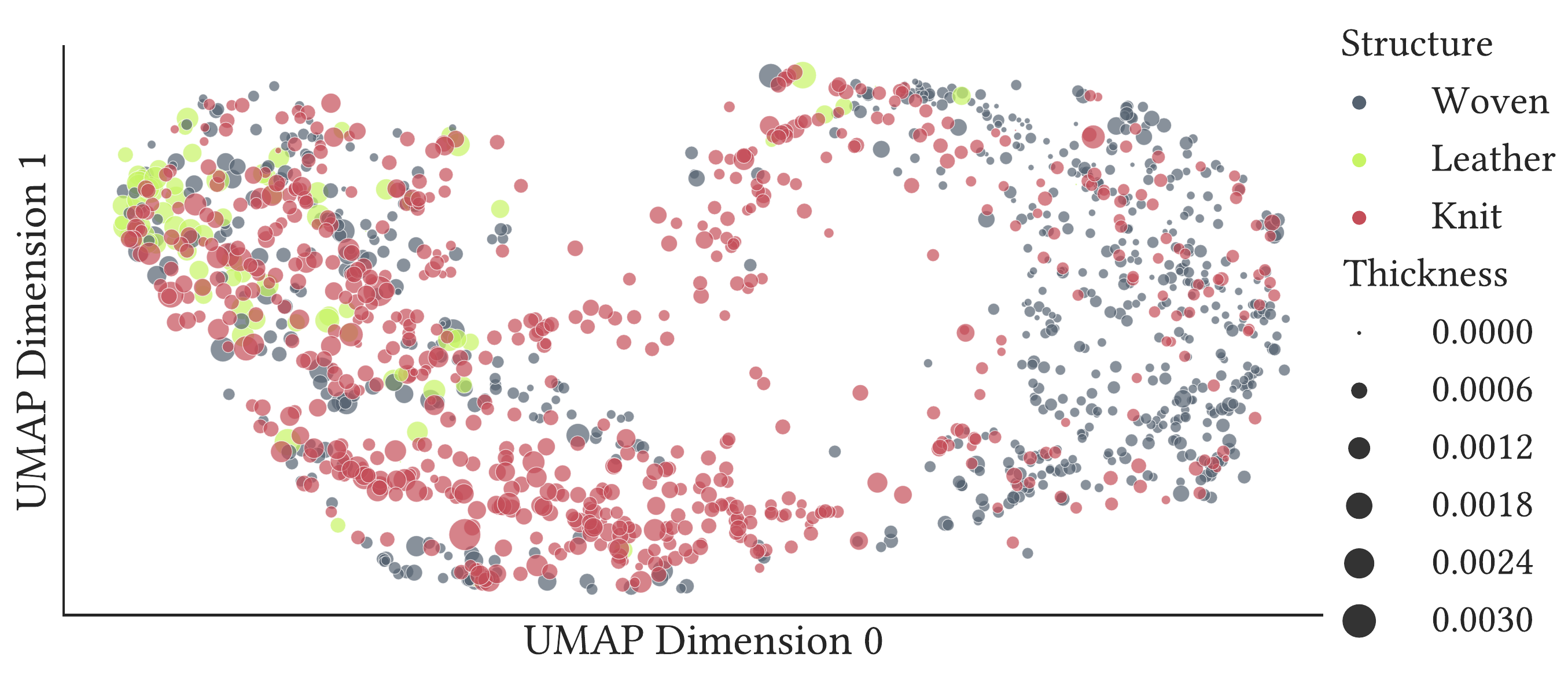}
	\vspace{-8mm}
	\caption{Embeddings of the transformer of our generator for data in the training set, reduced using UMAP~\cite{mcinnes2018umap}.
	}
	\vspace{-2mm}
	\label{fig:UMAP}
\end{figure}

We aim to understand which features are exploited by our model for making its predictions. In Figure~\ref{fig:UMAP}, we show the embeddings of our generator using UMAP. Despite some overlap, it seems that our model learns to separate between material \emph{families} (\emph{e.g.} \emph{leathers} from \emph{wovens}). Interestingly, the \emph{thickness} of the material is also a relevant parameter. The model is exploiting these semantic patterns without explicit supervision, providing evidence that material microgeometry plays important an important role for its optical appearance.

\subsection{Uncertainty Evaluation}\label{sec:uncertainty}

\REMOVE{
	\begin{figure}[tb!]
		\centering
		\includegraphics[width=\columnwidth]{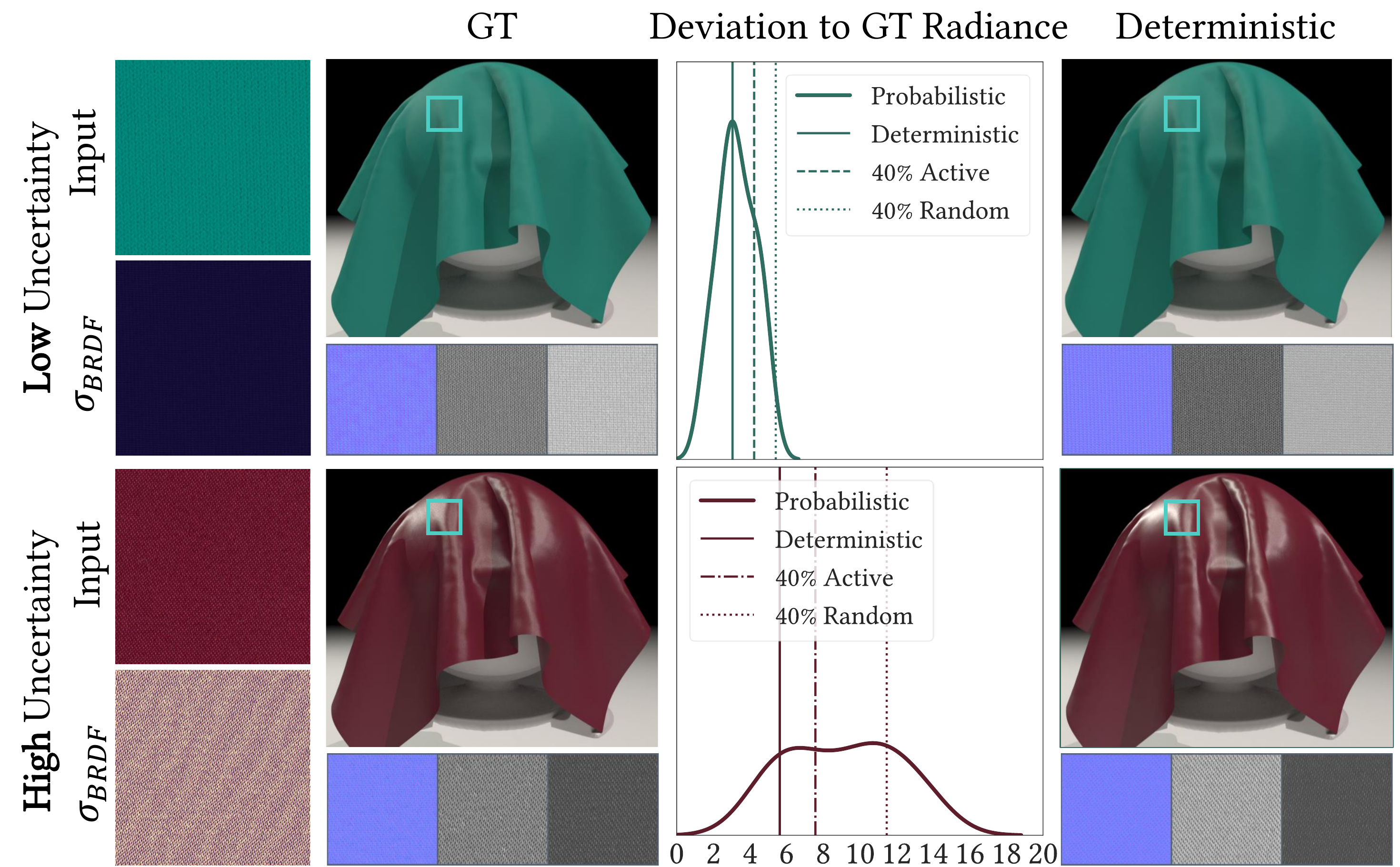}
		\vspace{-5mm}
		\caption{Examples of a material with low uncertainty $\sigma_{BRDF}$ (top row) and a material with high $\sigma_{BRDF}$ (bottom). For each, we include the input image, the ground truth maps, the estimations of our model on its deterministic configuration and three probabilistic estimations. As shown, the material with high uncertainty exhibits a more varied behaviour in terms of overall specularity.
		}
		
		\label{fig:uncertainty_high_low}
	\end{figure}
}

\REMOVE{
	\begin{figure*}
		\centering
		\begin{subfigure}{.35\textwidth}
			\centering
			\includegraphics[width=\linewidth]{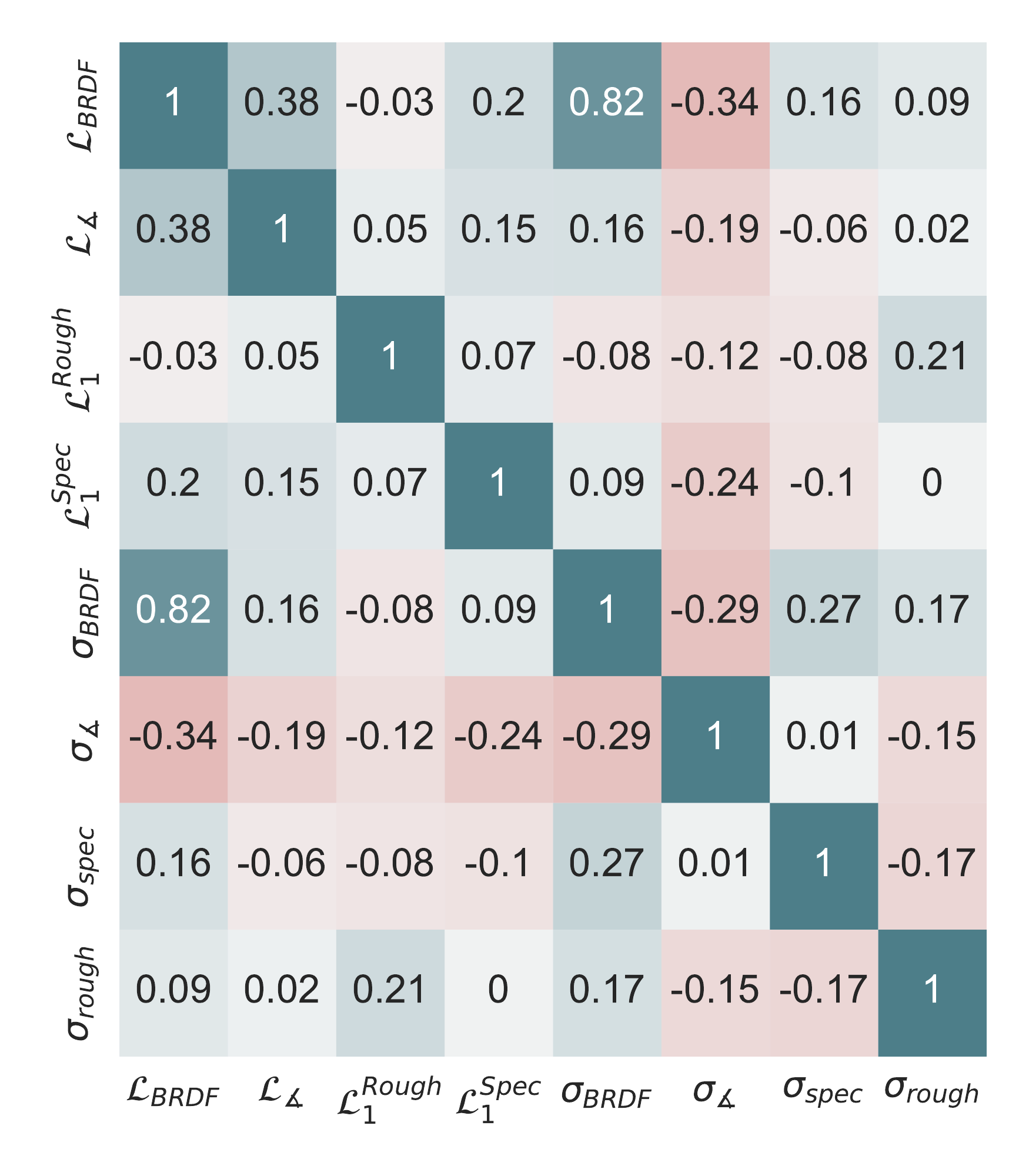}
			\label{fig:corr_matrix}
		\end{subfigure}
		\begin{subfigure}{.64\textwidth}
			\centering
			\vspace{-13mm}
			\includegraphics[width=\linewidth]{figures/pdf/uncertainty.pdf}
			
			\label{fig:uncertainty}
		\end{subfigure}
		
		\vspace{-8mm}
		\caption{On the left, we show a Pearson correlation matrix between per-map and render errors ($\mathcal{L}$) and uncertainties ($\sigma$) measured in the test set. On the right, for an \emph{rib} material with sequins, predicted maps (top row) and render and per-map uncertainties.}
		\label{fig:uncertainty_big}
	\end{figure*}
}
\REMOVE{
	\begin{figure}[tb!]
		\centering
		\vspace{-3mm}
		\includegraphics[width=\columnwidth]{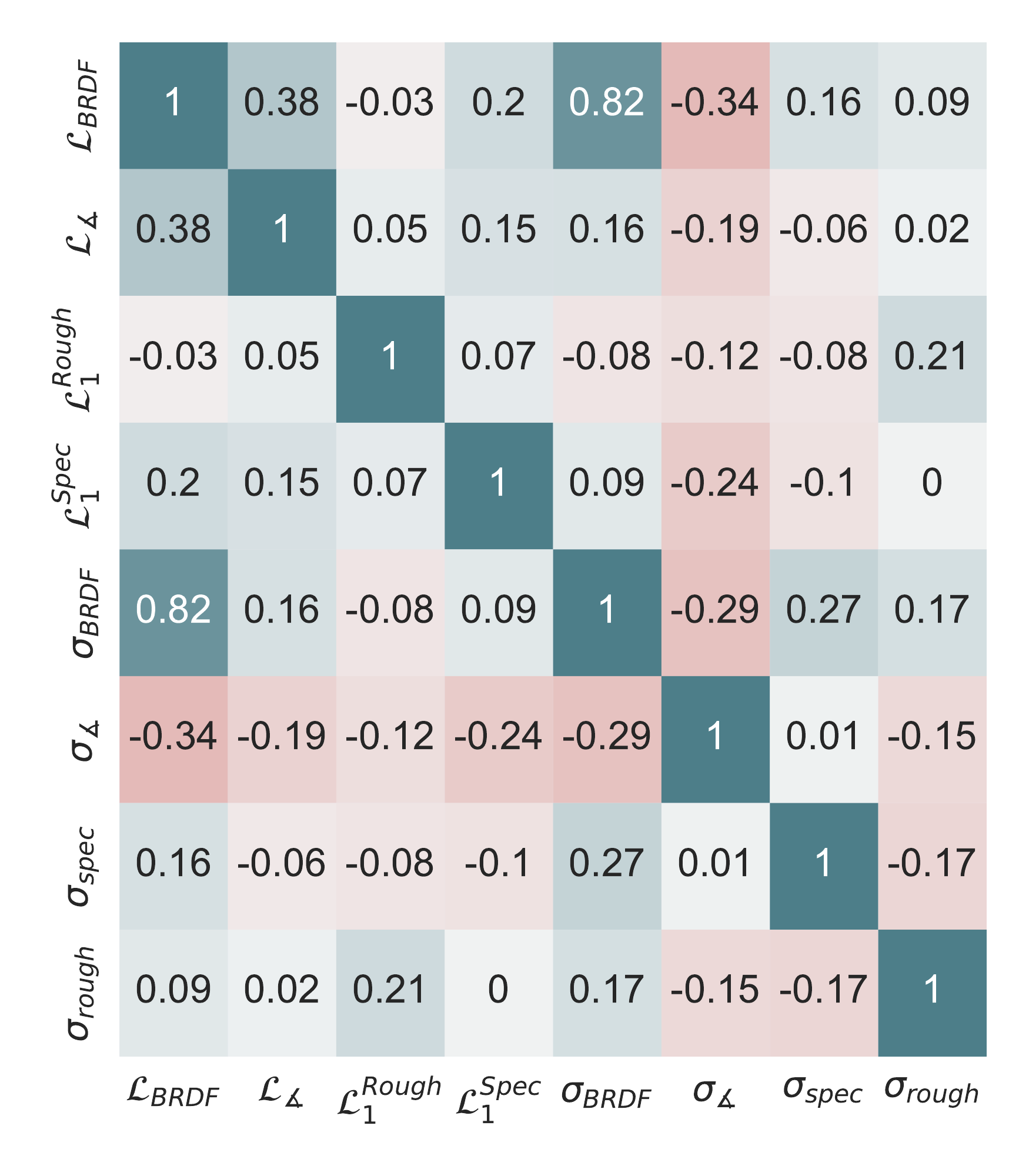}\caption{Pearson correlation matrix between per-map and render errors ($\mathcal{L}$) and uncertainties ($\sigma$) measured in the test set}
		\label{fig:corr_matrix}
\end{figure}}

\begin{figure}[tb!]
	\centering
	\vspace{-0mm}
	\includegraphics[width=\linewidth]{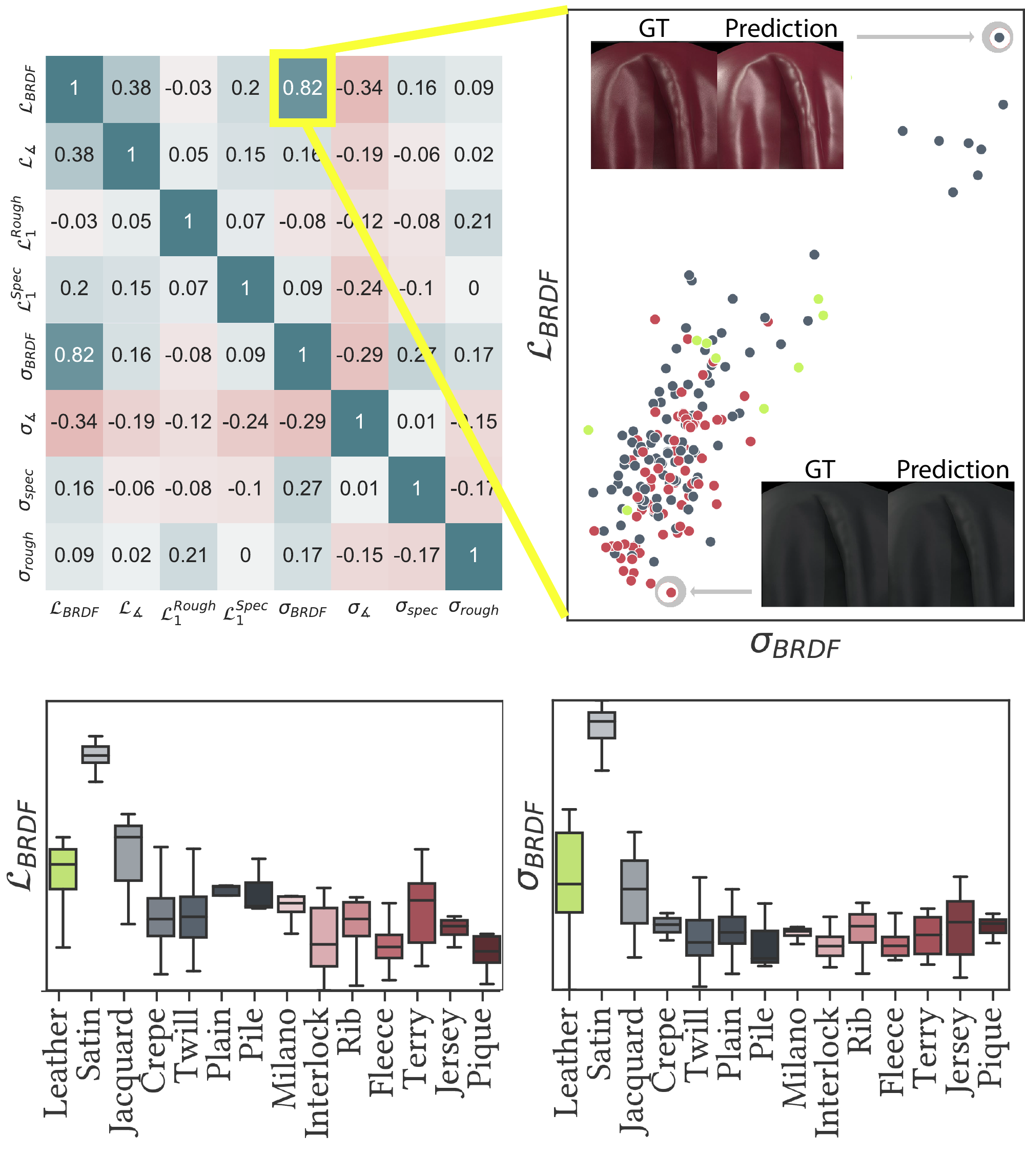}
	\vspace{-7mm}
	\caption{Top-left, correlation between render and pixel-wise losses, and uncertainties. Top-right, plot showing the correlation between our uncertainty metric and the error in render space; the renders illustrate the worst and best cases. Bottom, uncertainty and errors for different material families of the test set.
	}
	\vspace{-6mm}
	\label{fig:quantitative_analysis}
\end{figure}

In Figure~\ref{fig:quantitative_analysis}, we show the Pearson correlation matrix between the uncertainty and error for each map, our uncertainty metric (Equation~\ref{eq:uncertainty}), and the render perceptual metric (Equation~\ref{eq:brdf}), on our test set. As shown, neither the error nor uncertainties per map explain errors on the render space.  Our proposed $\umetric$ achieves a remarkable correlation of $82\%$ with the render error, validating that this metric is useful to predict errors at test time with reasonably high precision. 
In the same plot at the bottom, we distill the uncertainty and render error per family, in which we observe that our model struggles to accurately and confidently predict the reflectance of \emph{satins}, \emph{jacquards}, and \emph{leathers} more than it does for any other structure in our dataset. These structures have complex optical behaviour that make their digitization more challenging (for example, satins exhibit anisotropy, which we do not support in our material model), and are relatively uncommon in our training dataset. %
Figures~\ref{fig:teaser} and \ref{fig:uncertainty} show further examples of our uncertainty estimation for a diverse set of materials.

\REMOVE{
	\begin{figure}[tb!]
		\centering
		\vspace{-3mm}
		\includegraphics[width=\columnwidth]{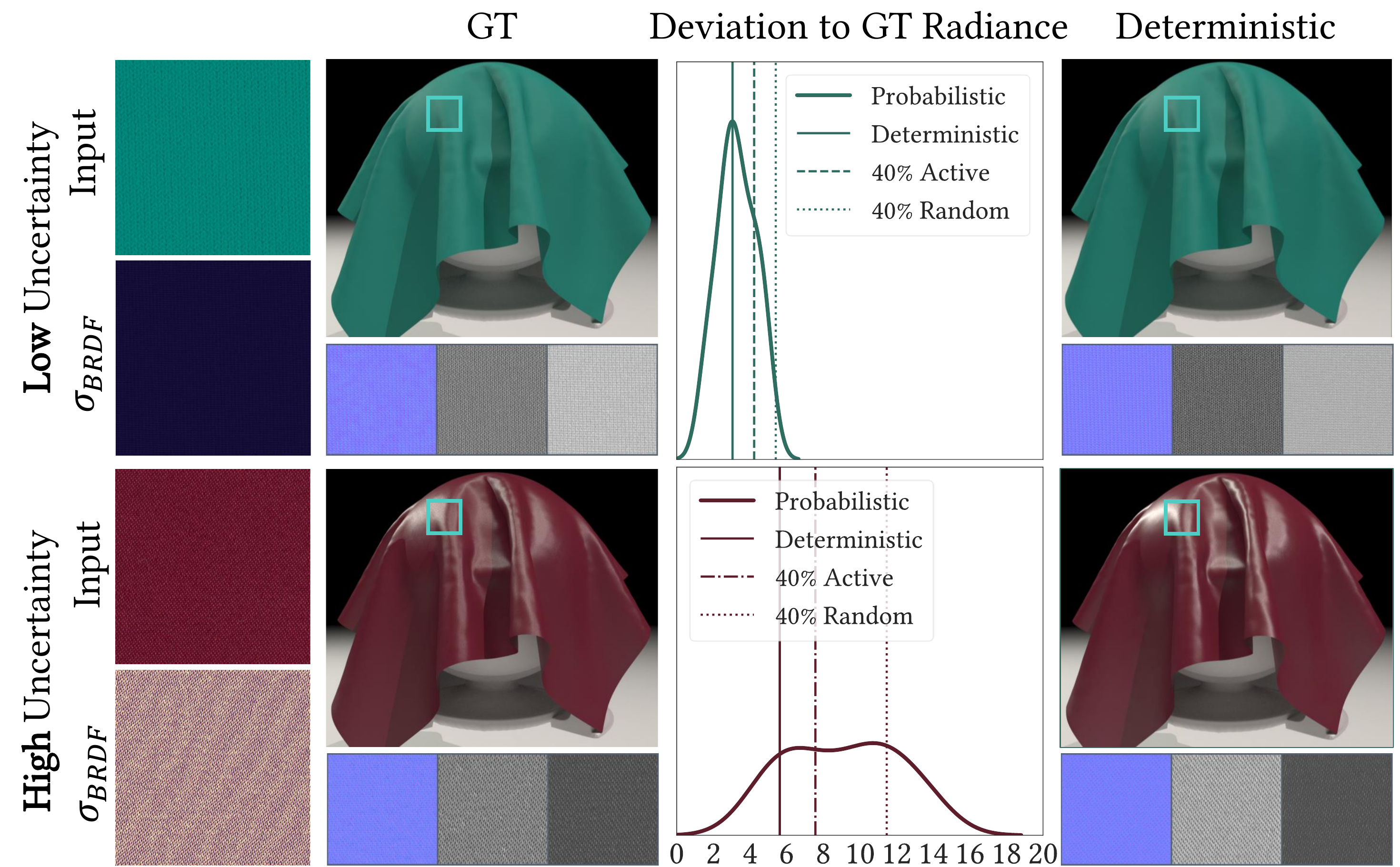}\caption{Examples of a material with low uncertainty $\umetric$ (top row) and a material with high $\umetric$ (bottom). For each, we include the input image, the ground truth maps, the estimations of our model on its deterministic configuration and three probabilistic estimations. As shown, the material with high uncertainty exhibits a more varied behaviour in terms of overall specularity.}
		\label{fig:uncertainty_high_low}
	\end{figure}
}

\subsection{Active Learning}\label{sec:activelearning}

\begin{figure}[tb!]
	\centering
	\includegraphics[width=\columnwidth]{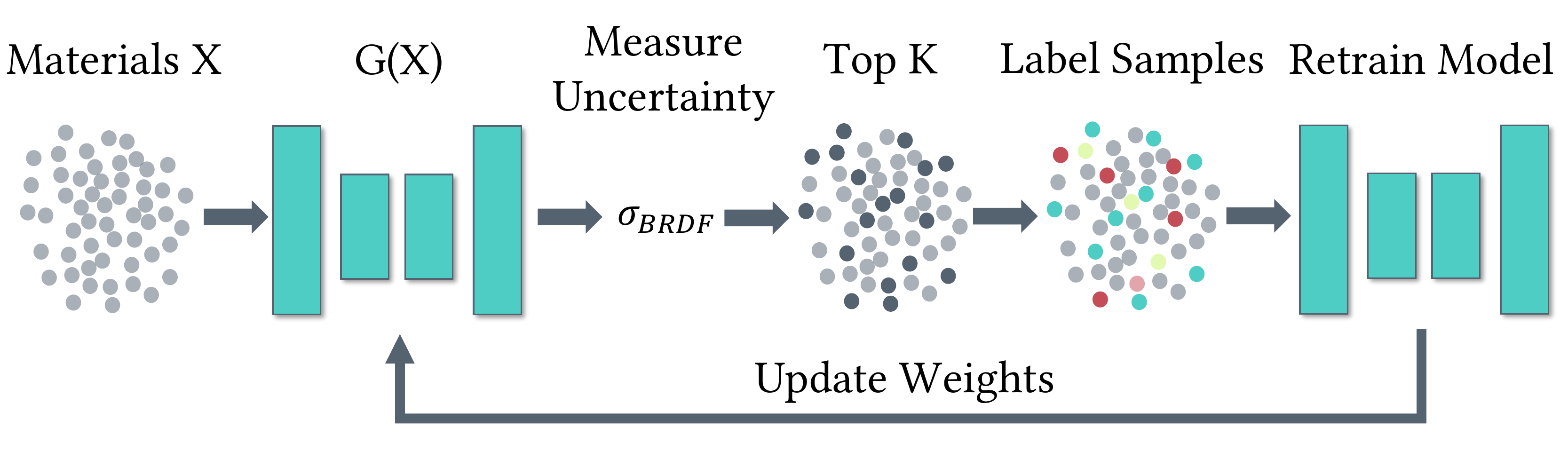}
	\vspace{-1mm}
	\includegraphics[width=\columnwidth]{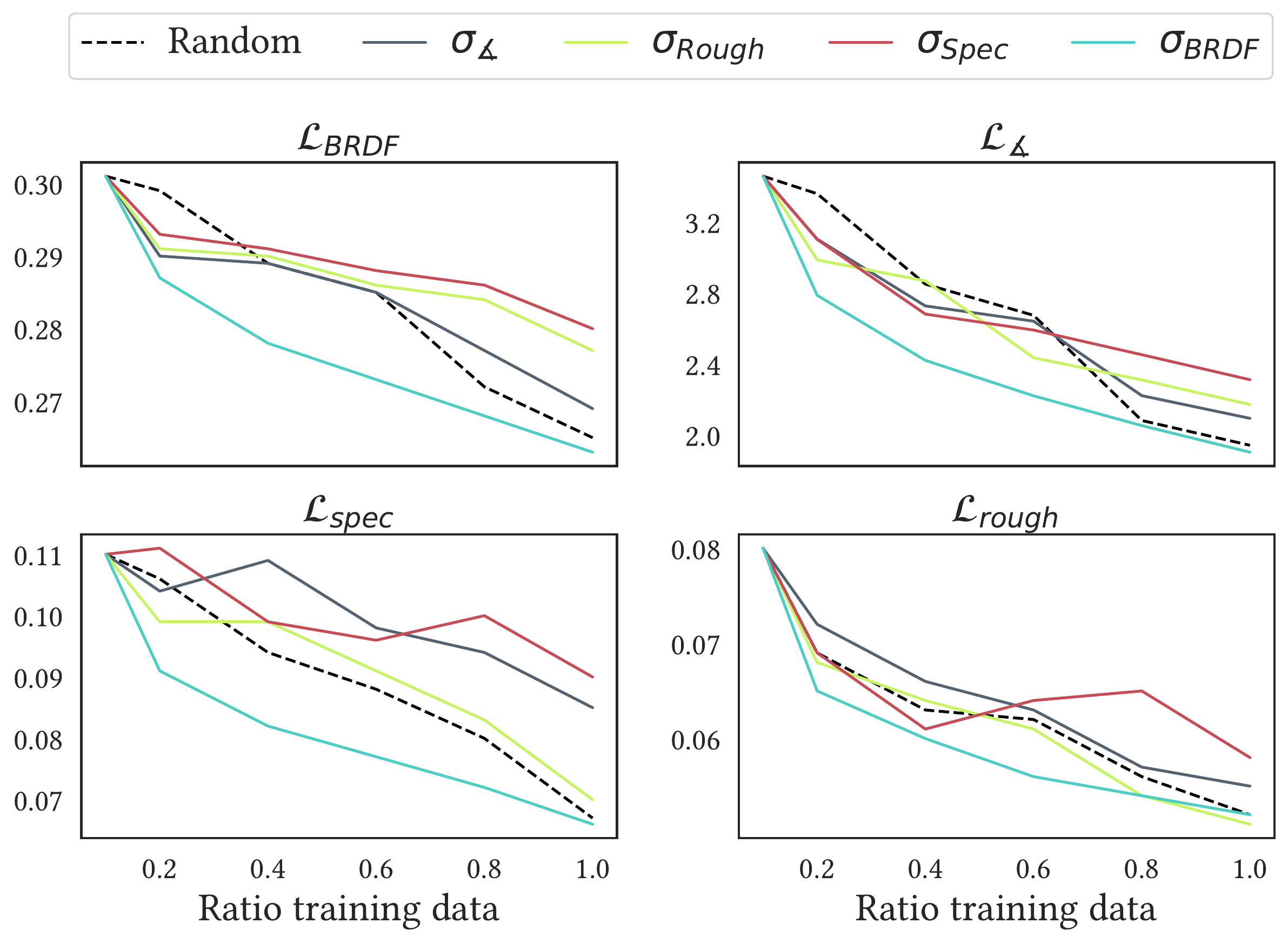}
	\vspace{-6mm}
	\caption{On top, illustration of our active learning algorithm. \REMOVE{From a set of materials, we measure the uncertainty in the render space for each, and select the \emph{top-k} samples with the highest uncertainty. We label those, aggregate them to the training dataset and retrain the model.} On the bottom, results of our active learning experiments. Leveraging $\umetric$ for actively selecting the top-k samples with the highest uncertainty, we achieve better results than a random sampling strategy for every metric we measure.}
	\vspace{-6mm}
	\label{fig:active_results}
\end{figure}

We leverage $\umetric$ for active learning~\cite{soleimany2021evidential} to identify the samples that contribute most to reduce the error of the model. Figure~\ref{fig:active_results}~(top) illustrates the process.
We start by training a model with 10\% of the available training data and measure the uncertainties in the remainder of the dataset. %
We then select the samples with the highest uncertainties and re-train the model with 20\%. We repeat the process with 40, 60, 80 and 100\% of the training dataset.
For each subset, we compare the performance of this model with four baselines: random-sampling, sampling by the highest uncertainty in normals, roughness, and specular. The results are shown in Figure~\ref{fig:active_results} (bottom). Active sampling based on $\umetric$ provides significant gains on sample efficiency, obtaining better accuracy for every map. 
For instance, an actively trained model with our metric which uses $20\%$ of the training data obtains comparable results to a model trained on three times more (but randomly sampled) data. 
While $\unormals$ is typically more informative than $\urough$ and $\uspec$, using per-map uncertainties does not provide better results than a random strategy. 
Finally, Figure~\ref{fig:teaser} shows the variation of the probabilistic samples with respect to the ground truth radiance for two materials with high and low uncertainty.

\subsection{Comparisons with Previous Work}\label{sec:comparisons}

In Table~\ref{tab:comparisons}, we compare our method with previous work on single image material capture. First, we made sure that the training data for these methods included textile materials similar to the ones we choose for testing. Emulating their capture conditions, we took images with a smartphone, with flash and ambient lighting. 
Note that this capture conditions are not ideal for our method, affecting the final renders if the albedo has shading gradients.
However, our goal in this experiment is to evaluate the overall preservation of the material structure in the inferred maps, particularly visible in the normals. 
For Shi~\etal~\cite{shi2020match}, we initialize the graph using a fabric material, provided in their open-source implementation and include the metallic map. Our model provides sharper and more accurate estimations without requiring optimization during test. Methods trained on style losses~\cite{henzler2021neuralmaterial, shi2020match} degrade the semantic structure, while Zhou~\etal~\cite{zhou2021adversarial} generate similar albedos to ours (note that ours are captured), but provide over-smooth estimations. We provide a comparison of timings and model sizes in Table~\ref{tab:computational-cost}. With our efficient model design, we can provide real time estimations without needing any optimization, which also enables our sampling-based uncertainty metric. Besides, it can also handle larger resolutions than previous work. Further results and re-rendered images are included in the supplementary material.

\begin{table}[]
	\centering
	\resizebox{\columnwidth}{!}{%
		\begin{tabular}{@{}rccc@{}}
			\toprule
			\multicolumn{1}{c}{\textbf{Method}}              & \textbf{Size (MB)}           & \textbf{Output Dims} & \textbf{Eval Time (s)} \\ \midrule
			Deep Inverse Rendering\textbf{*}~\cite{gao2019deep}                & 167                       & 256x256     & 603.5                  \\ \midrule
			Generative Modeling\textbf{*}~\cite{henzler2021neuralmaterial}                   & 1095.3                   & 512x384     &  218.8                 \\ \midrule
			Diff. Material Graphs\textbf{*}~\cite{shi2020match}                  & -                         & 512x512     &         1209.8            \\ \midrule
			Adversarial Estimation~\cite{zhou2021adversarial} & {11552.2} & 256x256     &    0.078                 \\\midrule
			\multirow{2}{*}{\textbf{UMat (Ours)}}             & \multirow{2}{*}{\GreenColor{\textbf{22.6}}}         & 256x256     &  \GreenColor{\textbf{0.036}}                    \\
			&                           & 512x512     &  \GreenColor{\textbf{0.131}} \\  
			\bottomrule                
		\end{tabular}%
	}
\vspace{-3mm}
	\caption{Model sizes for different methods, and their evaluation time in seconds (average of 100 evaluations on an RTX 2080 GPU), for different output sizes. The methods with \textbf{*} perform test-time optimization. DiffMat~\cite{shi2020match} does not use a pre-trained model.}
 	\vspace{-5mm}
	\label{tab:computational-cost}
\end{table}

\REMOVE{
\begin{table}[]
	
	\centering
	\resizebox{\columnwidth}{!}{%
		\begin{tabular}{@{}rcccccc@{}}
			\toprule
			\multicolumn{1}{c}{\textbf{Method}} &~\cite{gao2019deep}  & ~\cite{henzler2021neuralmaterial} &~\cite{shi2020match}      & ~\cite{zhou2021adversarial} & \multicolumn{2}{c}{\textbf{UMat}} \\ \midrule
			\textbf{Model Size (MB)}        & 167     & 1095.29 & -       & 11552.22 & \multicolumn{2}{c}{\GreenColor{\textbf{22.6}}} \\ \midrule
			\textbf{Output Dims}       & 256x256 & 512x384 & 512x512 & 256x256  & 256x256   & 512x512  \\ \midrule
			\textbf{Eval Time (s)} &  603.5       &   218.8       &    1209.8       &   0.078            &   \GreenColor{\textbf{0.036}}                &    \GreenColor{\textbf{0.131}}      \\ \bottomrule
		\end{tabular}%
	}
	\caption{}
	\label{tab:my-table}
\end{table}
}

\newcommand{\addpic}{\includegraphics[width=0.145\textwidth]{example-image}}
\newcolumntype{C}{>{\centering\arraybackslash}m{.145\textwidth}}
\begin{table*}
	\centering
	\vspace{-3mm}
	\begin{tabular}{C*5{C}@{}}
		\toprule
		Input & Deep Inverse Rendering~\cite{gao2019deep} & Generative Modeling~\cite{henzler2021neuralmaterial} & Diff. Material Graphs~\cite{shi2020match} & Adversarial Estimation~\cite{zhou2021adversarial} & \textbf{UMat (Ours)} \\ 
		\midrule
		\includegraphics[width=0.145\textwidth]{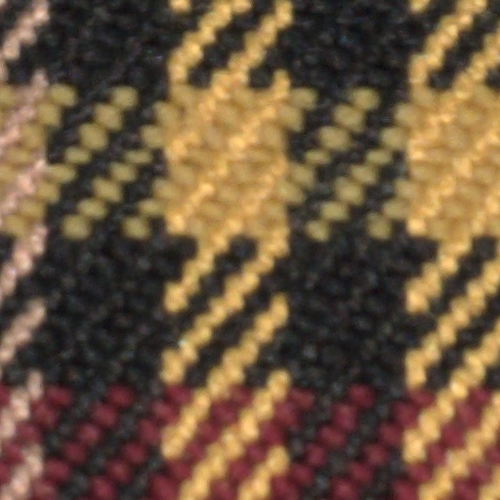}& \includegraphics[width=0.145\textwidth]{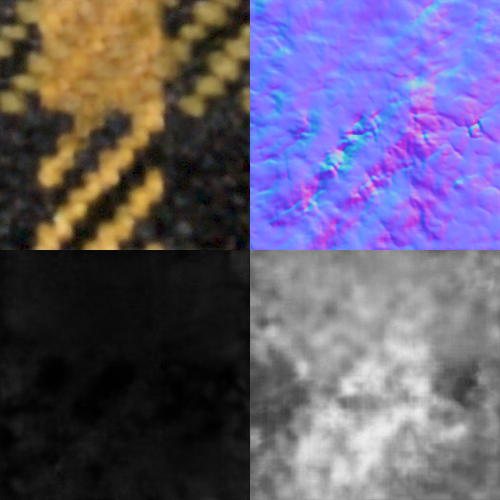} & \includegraphics[width=0.145\textwidth]{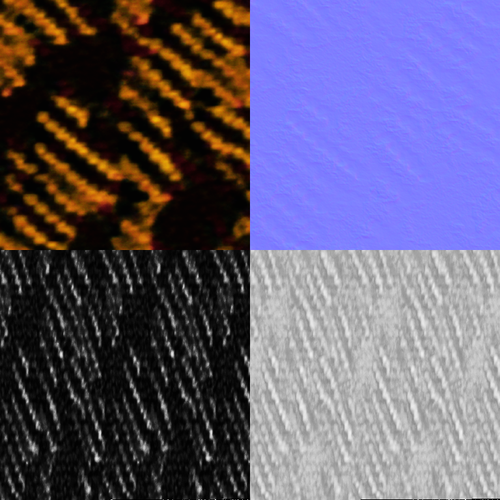} & \includegraphics[width=0.145\textwidth]{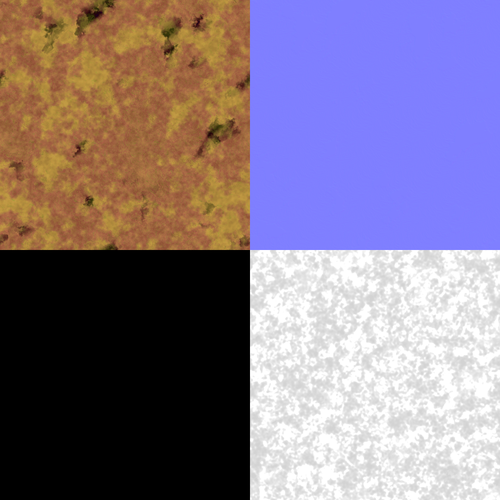} & \includegraphics[width=0.145\textwidth]{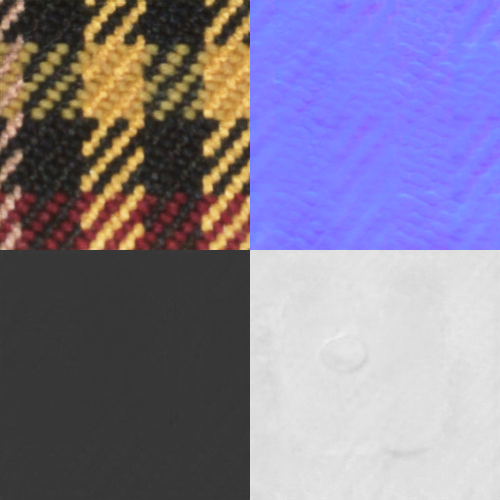}  & \includegraphics[width=0.145\textwidth]{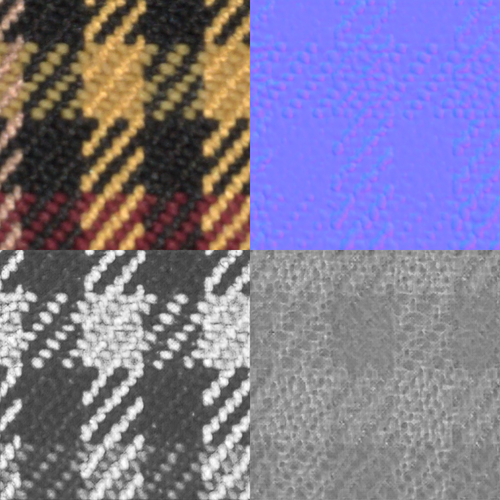}   \\ 
		\includegraphics[width=0.145\textwidth]{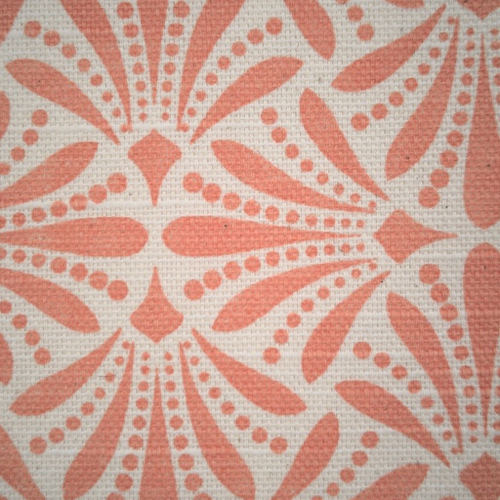}& \includegraphics[width=0.145\textwidth]{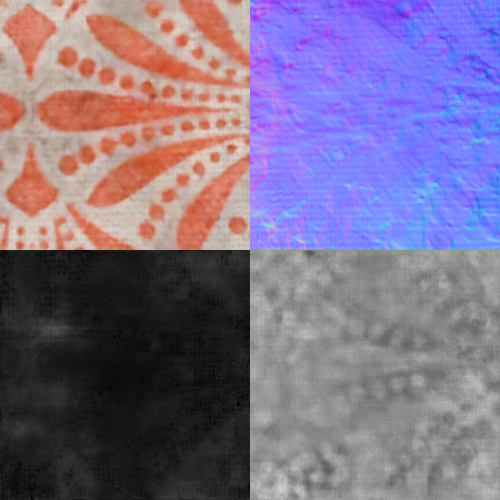} & \includegraphics[width=0.145\textwidth]{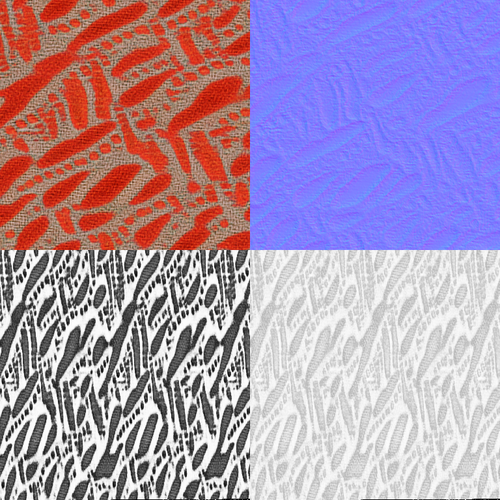} & \includegraphics[width=0.145\textwidth]{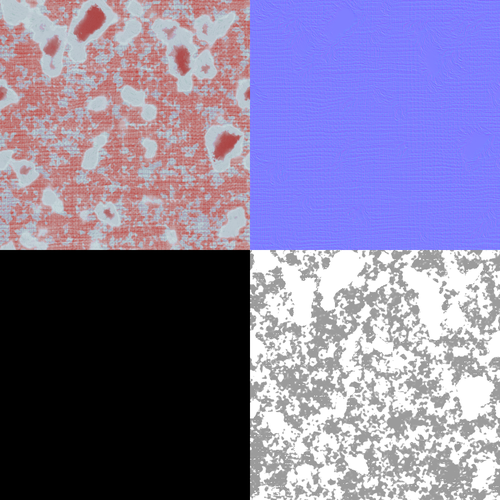} & \includegraphics[width=0.145\textwidth]{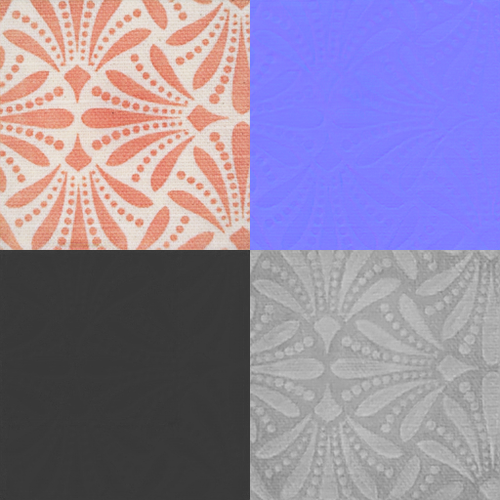}  & \includegraphics[width=0.145\textwidth]{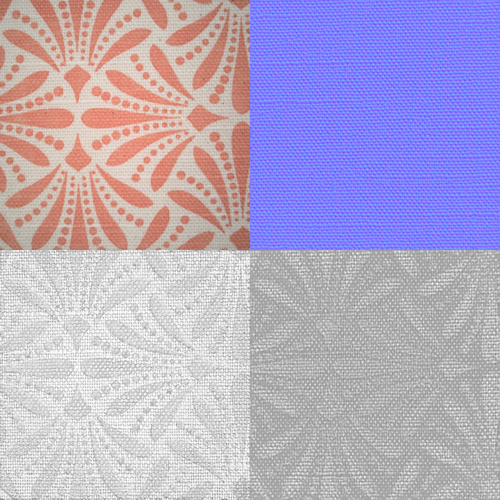}   \\ 
		\includegraphics[width=0.145\textwidth]{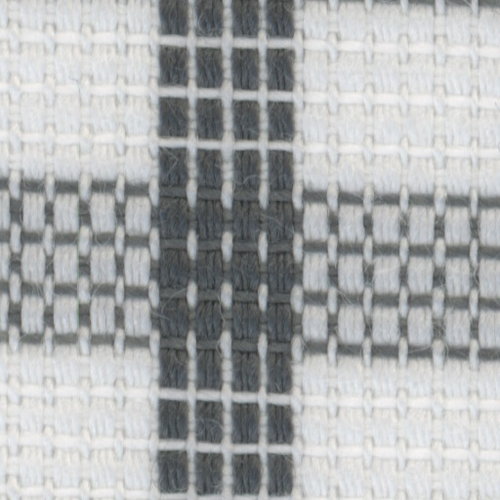}& \includegraphics[width=0.145\textwidth]{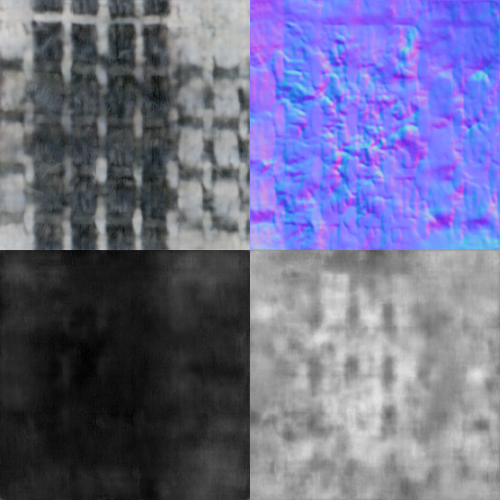} & \includegraphics[width=0.145\textwidth]{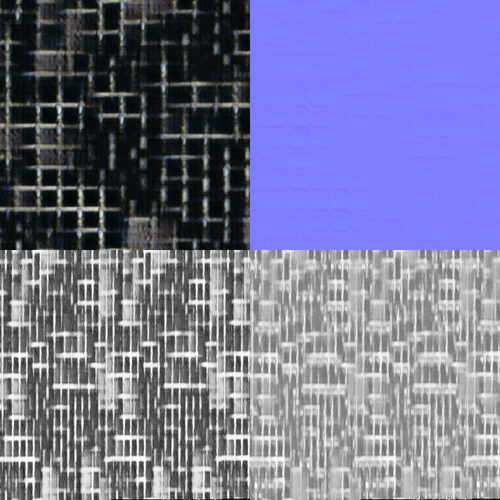} & \includegraphics[width=0.145\textwidth]{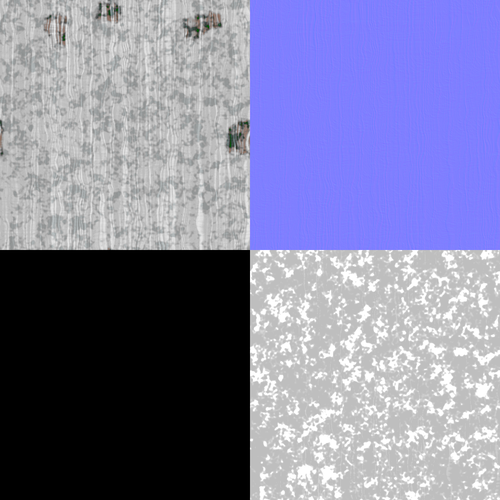} & \includegraphics[width=0.145\textwidth]{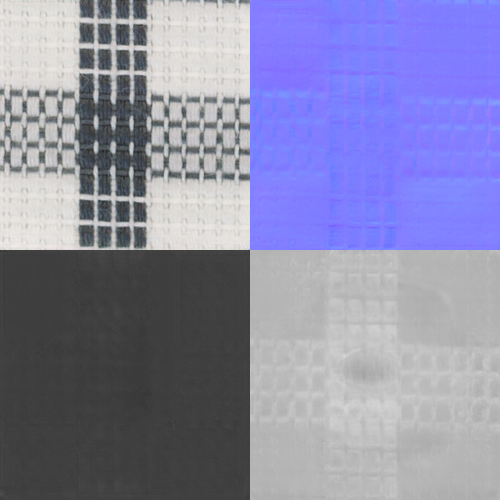}  & \includegraphics[width=0.145\textwidth]{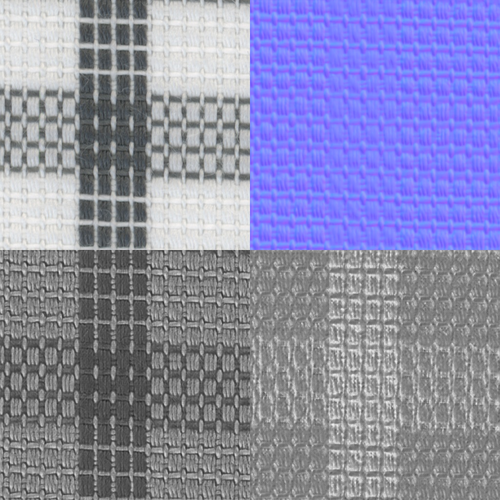}   \\ 
		\bottomrule 
	\end{tabular}
\vspace{-3mm}
	\caption{Comparisons of our method with previous work on images captured under different lighting conditions. Top: a smartphone flash-lit image. Middle: a smartphone image with ambient light. Bottom: a flatbed scanner capture. Our method produces the best results preserving the microstructure even when capture conditions degrade due to the sensor resolution. Note that we do not estimate albedos and that absolute intensities for specular and roughness maps are not directly comparable due to differences in the material model. }
	\vspace{-4mm}
	\label{tab:comparisons}
\end{table*}

\subsection{Limitations}
\begin{figure}[tb!]
	\centering
		\vspace{-2mm}
	\includegraphics[width=\columnwidth]{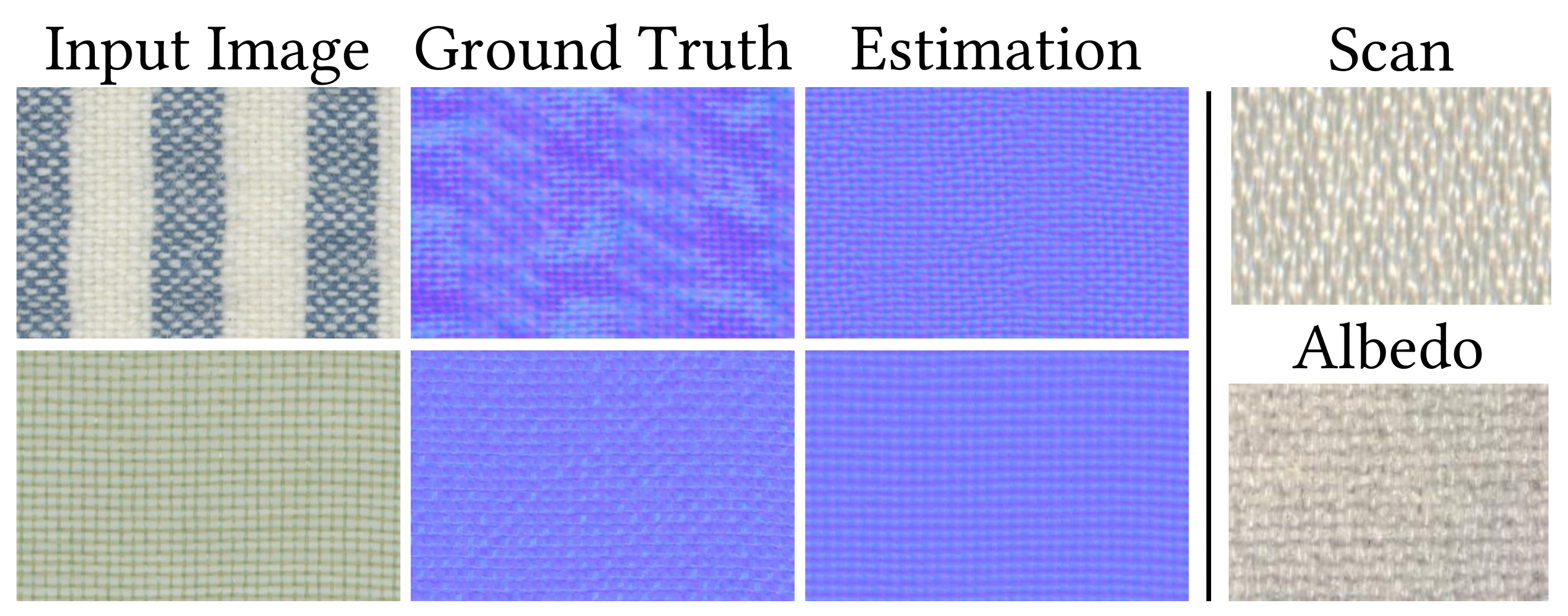}
	\vspace{-6mm}
	\caption{Limitations cases of our method. On the left, we show a \emph{seersucker}, with wrinkles that are hidden by the diffuse illumination of the device, and a translucent \emph{organza} with holes between the yarns that appear very bright due to the white background of the scanner, and are therefore mistakenly treated as yarn centers. On the right, we show that for highly directional materials, such as \emph{satins}, the diffuse-like illumination in our capture device sometimes introduces specular highlights.}
	\vspace{-6mm}
	\label{fig:failure_cases}
\end{figure}

We show some limitations in Figure~\ref{fig:failure_cases}. The illumination in the scanner hides the wrinkles in the \emph{seersucker} fabric, and our model predicts a flat surface. 
The \emph{organza} fabric at the bottom is very transparent with visible holes between the yarns. Since the background is white, the model has mistakenly treated the light regions as yarn centers. For the \emph{satin} at the right, the scan image exhibits specular highlights due to the directionality of the yarns. While this image may be problematic to use as an albedo, it does not affect our metrics as we use constant albedos to compute them.

\section{Conclusions}\label{sec:conclusions}

We have presented a GAN-based method to digitize materials which leverages microgeometry appearance and a flatbed scanner as capture device.
Our method has shown better performance than state-of-the-art solutions that require a single image as input, when it comes to textile materials. 
To account for potential ambiguities derived from the capture setting, we have presented a method to model the uncertainty in the estimation at test time. 

Managing uncertainty in machine learning projects is important to guarantee robust and functional solutions. However, this typically comes at the cost of complex or slow models. In this work, we have presented the first method to quantify uncertainty in single image material digitization, while introducing minimal impact in the training and evaluation processes. 
While it is currently not possible to discern the source of the uncertainty, whether it is epistemic (uncertainty which can be reduced by increasing the dataset size) or aleatoric (which is derived by a noisy data generation process), our metric has proven useful to identify ambiguous inputs, underrepresented classes, or out-of-distribution data.

We could extend our work in several ways. The most obvious extension is to estimate real albedos, so that we can deal with other type of scanning devices. Further, expanding our material model to give support for more reflectance properties, such as transmittance or anisotropy, could be useful to improve the realism in render of textiles. 
Finally, we will keep growing our dataset according to our active sampling process to add more families.

{\small
\paragraph*{Acknowledgments}~Elena Garces was partially supported by a Juan de la Cierva - Incorporacion Fellowship (IJC2020-044192-I).We thank Jorge Lopez-Moreno and Dan Casas for valuable discussions, and Sofía Domínguez for helping build the dataset. 
}
\vspace{-5mm}
{\small
\bibliographystyle{ieee_fullname}
\bibliography{egbib}
}

\end{document}